\documentclass[twoside]{article}

\usepackage[accepted]{aistats2021}
\usepackage[round]{natbib}

\usepackage[utf8]{inputenc} 
\usepackage[T1]{fontenc}    
\usepackage{hyperref}       
\usepackage{url}            
\usepackage{booktabs}       
\usepackage{amsfonts}       
\usepackage{nicefrac}       
\usepackage{microtype}      
\usepackage{microtype}
\usepackage{graphicx}
\usepackage{subfigure}
\usepackage{booktabs} 
\usepackage{algorithmic}
\usepackage{algorithm}
\usepackage{color}
\usepackage{mathrsfs}
\usepackage{float}
\usepackage{amsmath} 
\usepackage{amssymb}
\usepackage{amsfonts}
\usepackage{amsthm}
\usepackage{stmaryrd}
\usepackage{stackrel}
\usepackage{graphicx}
\usepackage{hyperref}
\usepackage{dsfont}
\usepackage{tikz}
\usepackage{framed}
\usepackage{appendix}
\usepackage{cleveref}
\usepackage{enumerate}
\usepackage{placeins}
\usepackage{authblk}
\usepackage{textcomp}
\usepackage{lipsum,adjustbox}
\usepackage{caption}
\usepackage{colortbl}

\newcommand{\Dn}{\mathscr{D}_{n}}
\newcommand{\bX}{\textbf{X}}

\newcommand{\bx}{\textbf{x}}
\renewcommand{\P}{\mathds{P}}
\newcommand{\R}{\mathds{R}}
\newcommand{\E}{\mathbb{E}}

\renewcommand{\path}{\mathscr{P}}
\newcommand{\pathset}{\hat{\path}}

\newcommand{\setpstar}{\mathcal{U}^{\star}}

\numberwithin{equation}{section}
\theoremstyle{plain}
\newtheorem{theorem}{Theorem}

\begin{document}

%

%
\runningauthor{C. Bénard, G. Biau, S. Da Veiga, E. Scornet}

\twocolumn[

\aistatstitle{Interpretable Random Forests via Rule Extraction}

\aistatsauthor{ Clément Bénard$^{1,2}$ \And Gérard Biau$^{2}$ \And Sébastien Da Veiga$^{1}$ \And Erwan Scornet$^{3}$}

\aistatsaddress{ \\[-0.5em] $^{1}$Safran Tech, Modeling \& Simulation, 78114 Magny-Les-Hameaux, France \\ 
$^{2}$Sorbonne Université, CNRS, LPSM, 75005 Paris, France \\
$^{3}$Ecole Polytechnique, IP Paris, CMAP, 91128 Palaiseau, France } ]

\begin{abstract}
We introduce SIRUS (\textbf{S}table and \textbf{I}nterpretable \textbf{RU}le \textbf{S}et) for regression, a stable rule learning algorithm, which takes the form of a short and simple list of rules.
State-of-the-art learning algorithms are often referred to as ``black boxes'' because of the high number of operations involved in their prediction process. Despite their powerful predictivity, this lack of interpretability may be highly restrictive for applications with critical decisions at stake.
On the other hand, algorithms with a simple structure---typically decision trees, rule algorithms, or sparse linear models---are well known for their instability. This undesirable feature makes the conclusions of the data  analysis unreliable and turns out to be a strong operational limitation.
This motivates the design of SIRUS, based on random forests, which combines a simple structure, a remarkable stable behavior when data is perturbed, and an accuracy comparable to its competitors.
We demonstrate the efficiency of the method both empirically (through experiments) and theoretically (with the proof of its asymptotic stability). A \texttt{R/C++} software implementation \texttt{sirus} is available from \texttt{CRAN}.
\end{abstract}

\section{Introduction}

State-of-the-art learning algorithms, such as random forests or neural networks, are often criticized for their ``black-box" nature. This criticism essentially results from the high number of operations involved in their prediction mechanism, as it prevents to grasp how inputs are combined to generate predictions. Interpretability of machine learning algorithms is receiving an increasing amount of attention since the lack of transparency is a strong limitation for many applications, in particular those involving critical decisions. 
The analysis of production processes in the manufacturing industry typically falls into this category. Indeed, such processes involve complex physical and chemical phenomena that can often be successfully modeled by black-box learning algorithms. However, any modification of a production process has deep and long-term consequences, and therefore cannot simply result from a blind stochastic modelling. In this domain, algorithms have to be interpretable, i.e., provide a sound understanding of the relation between inputs and outputs, in order to leverage insights to guide physical analysis and improve efficiency of the production.

Although there is no agreement in the machine learning litterature about a precise definition of interpretability \citep{lipton2016mythos, murdoch2019interpretable}, it is yet possible to define simplicity, stability, and predictivity as minimum requirements for interpretable models \citep{benard2020sirus,yu2019three}. Simplicity of the model structure can be assessed by the number of operations performed in the prediction mechanism. In particular, \citet{murdoch2019interpretable} introduce the notion of \textit{simulatable models} when a human is able to reproduce the prediction process by hand. Secondly, \citet{yu2013stability} argues that ``interpretability needs stability'', as the conclusions of a statistical analysis have to be robust to small data perturbations to be meaningful. Instability is the symptom of a partial and arbitrary modelling of the data, also known as the \textit{Rashomon effect} \citep{breiman2001statistical}. Finally, as also explained in \citet{breiman2001statistical}, if the decrease of predictive accuracy is significant compared to a state-of-the-art black-box algorithm, the interpretable model misses some patterns in the data and is therefore misleading.

Decision trees \citep{leo1984classification} can model nonlinear patterns while having a simple structure. They are therefore often presented as interpretable. However, the structure of trees is highly sensitive to small data perturbation \citep{breiman2001statistical}, which violates the stability principle and is thus a strong limitation to their practical use. Rule algorithms are another type of nonlinear methods with a simple structure, defined as a collection of elementary rules. An elementary rule is a set of constraints on input variables, which forms a hyperrectangle in the input space and on which the associated prediction is constant. As an example, such a rule typically takes the following simple form: 
\begin{center}
	\setlength{\fboxrule}{1pt}
	\fbox{\begin{minipage}{0.46 \textwidth}
			\textbf{If} \hspace{0.5mm} $\left\{ \hspace{-2mm} \begin{tabular}{c} $X^{(1)} < 1.12$ \\ \textbf{\&} $X^{(3)} \geq 0.7$ \end{tabular}\right.$ \textbf{then} $\hat{Y} = 0.18$ \textbf{else} $\hat{Y} = 4.1$ \textbf{.}
	\end{minipage}}
\end{center}
A large number of rule algorithms have been developed, among which the most influential are Decision List \citep{rivest1987learning}, CN2 \citep{clark1989cn2}, C4.5 \citep{quinlan1992c4}, IREP \citep[Incremental Reduced Error Pruning,][]{furnkranz1994incremental}, RIPPER \citep[Repeated Incremental Pruning to Produce Error Reduction,][]{cohen1995fast}, 
PART \citep[Partial Decision Trees,][]{frank1998generating}, SLIPPER \citep[Simple Learner with Iterative Pruning to Produce Error Reduction,][]{cohen1999simple}, LRI \citep[Leightweight Rule Induction,][]{weiss2000lightweight}, RuleFit \citep{friedman2008predictive}, Node harvest \citep{meinshausen2010node},
ENDER \citep[Ensemble of Decision Rules,][]{dembczynski2010ender}, BRL \citep[Bayesian Rule Lists,][]{letham2015interpretable}, RIPE \citep[Rule Induction Partitioning Estimator,][]{margot2018rule, margot2019consistent}, and
\citet[Generalized Linear Rule Models]{wei2019generalized}.
It turns out, however, that despite their simplicity and high predictivity (close to the accuracy of tree ensembles), rule learning algorithms share the same limitation as decision trees: instability.
Furthermore, among the hundreds of existing rule algorithms, most of them are designed for supervised classification and few have the ability to handle regression problems.

The purpose of this article is to propose a new stable rule algorithm for regression, SIRUS (\textbf{S}table and \textbf{I}nterpretable \textbf{RU}le \textbf{S}et), and therefore demonstrate that rule methods can address regression problems efficiently while producing compact and stable list of rules. To this aim, we build on \citet{benard2020sirus}, who have introduced SIRUS for classification problems. Our algorithm is based on random forests \citep{breiman2001random}, and its general principle is as follows: since each node of each tree of a random forest can be turned into an elementary rule, the core idea is to extract rules from a tree ensemble based on their frequency of appearance. The most frequent rules, which represent robust and strong patterns in the data, are ultimately linearly combined to form predictions.
The main competitors of SIRUS are RuleFit \citep{friedman2008predictive} and Node harvest \citep{meinshausen2010node}. Both methods also extract large collection of rules from tree ensembles: RuleFit uses a boosted tree ensemble \citep[ISLE,][]{friedman2003importance} whereas Node harvest is based on random forests. The rule selection is performed by a sparse linear aggregation, respectively the Lasso \citep{tibshirani1996regression} for RuleFit and a constrained quadratic program for Node harvest. Yet, despite their powerful predictive skills, these two methods tend to produce long, complex, and unstable lists of rules (typically of the order of $30-50$), which makes their interpretability questionable. Because of the randomness in the tree ensemble, running these algorithms multiple times on the same dataset outputs different rule lists.
As we will see, SIRUS considerably improves stability and simplicity over its competitors, while preserving a comparable predictive accuracy and computational complexity---see Section $2$ of the Supplementary Material for the complexity analysis.

We present SIRUS algorithm in Section \ref{sec_sirus}. In Section \ref{sec_xp}, experiments illustrate the good performance of our algorithm in various settings. Section \ref{sec_theory} is devoted to studying the theoretical properties of the method, with, in particular, a proof of its asymptotic stability. Finally, Section \ref{conclusion} summarizes the main results and discusses research directions for future work. 
Additional details are gathered in the Supplementary Material. 

\section{SIRUS Algorithm} \label{sec_sirus}

We consider a standard regression setting where we observe an i.i.d.~sample $\mathscr{D}_{n} = \{(\bX_{i},Y_{i}), i=1, \hdots, n\}$, with each $(\bX_i,Y_i)$ distributed as a generic pair $(\bX,Y)$ independent of $\mathscr{D}_{n}$. The $p$-tuple $\bX=(X^{(1)},\hdots,X^{(p)})$ is a random vector taking values in $\mathds R^p$, and $Y \in \R$ is the response.
Our objective is to estimate the regression function $m(\bx) = \E [Y|\bX = \bx]$ with a small and stable set of rules.


\paragraph{Rule generation.}
The \textbf{first step} of SIRUS is to grow a random forest with a large number $M$ of trees based on the available sample $\Dn$. 
The critical feature of our approach to stabilize the forest structure is to restrict node splits to the $q$-empirical quantiles of the marginals $\smash{X^{(1)}, \hdots, X^{(p)}}$, with typically $q = 10$. This modification to Breiman's original algorithm has a small impact on predictive accuracy, but is essential for stability, as it is extensively discussed in Section $3$ of the Supplementary Material.
Next, the obtained forest is broken down in a large collection of rules in the following process.
First, observe that each node of each tree of the resulting ensemble defines a hyperrectangle in the input space $\R^p$. Such a node can therefore be turned into an elementary regression rule, by defining a piecewise constant estimate whose value only depends on whether the query point falls in the hyperrectangle or not. Formally, a (inner or terminal) node of the tree is represented by a path, say $\mathscr P$, which describes the sequence of splits to reach the node from the root of the tree. In the sequel, we denote by $\Pi$ the finite list of all possible paths, and insist that each path $\mathscr P \in \Pi$ defines a regression rule.
Based on this principle, in the first step of the algorithm, both internal and external nodes are extracted from the trees of the random forest to generate a large collection of rules, typically $10^4$.

\paragraph{Rule selection.}
The \textbf{second step} of SIRUS is to select the relevant rules from this large collection. Despite the tree randomization in the forest construction, there are some redundancy in the extracted rules.
Indeed those with a high frequency of appearance represent strong and robust patterns in the data, and are therefore good candidates to be  included in a compact, stable, and predictive rule ensemble. This occurrence frequency is denoted by $\hat{p}_{M,n}(\path)$ for each possible path $\path \in \Pi$. Then a threshold $p_0 \in (0,1)$ is simply used to select the relevant rules, that is 
\[
\pathset_{M,n,p_{0}}=\{ \path \in \Pi:\hat{p}_{M,n}(\path)>p_{0}\}.
\]
The threshold $p_0$ is a tuning parameter, whose influence and optimal setting are discussed and illustrated later in the experiments (Figures \ref{fig_diabetes} and \ref{fig_machine}). Optimal $p_0$ values essentially select rules made of one or two splits. Indeed, rules with a higher number of splits are more sensitive to data perturbation, and thus associated to smaller values of $\hat{p}_{M,n}(\path)$. Therefore, SIRUS grows shallow trees to reduce the computational cost while leaving the rule selection untouched---see Section $3$ of the Supplementary Material.
In a word, SIRUS uses the principle of randomized bagging, but aggregates the forest structure itself instead of predictions in order to stabilize the rule selection.

\paragraph{Rule set post-treatment.}
The rules associated with the set of distinct paths $\pathset_{M,n,p_{0}}$ are dependent by definition of the path extraction mechanism. 
As an example, let us consider the $6$ rules extracted from a random tree of depth $2$. Since the tree structure is recursive, $2$ rules are made of one split and $4$ rules of two splits. 
Those $6$ rules are linearly dependent because their associated hyperrectangles overlap.
Consequently, to properly settle a linear aggregation of the rules, the \textbf{third step} of SIRUS filters $\smash{\pathset_{M,n,p_{0}}}$ with the following post-treatment procedure: if the rule induced by the path $\smash{\path \in \pathset_{M,n,p_{0}}}$ is a linear combination of rules associated with paths with a higher frequency of appearance, then $\path$ is simply removed from $\smash{\pathset_{M,n,p_{0}}}$. We refer to Section $4$ of the Supplementary Material for a detailed illustration of the post-treatment procedure on real data.

\paragraph{Rule aggregation.}
By following the previous steps, we finally obtain a small set of regression rules.
As such, a rule $\hat{g}_{n,\path}$ associated with a path $\path$ is a piecewise constant estimate: if a query point $\bx$ falls into the corresponding hyperrectangle $H_{\path} \subset \R^p$, the rule returns the average of the $Y_i$'s for the training points $\bX_i$'s that belong to $H_{\path}$; symmetrically, if $\bx$ falls outside of $H_{\path}$, the average of the $Y_i$'s for training points outside of $H_{\path}$ is returned.
Next, a non-negative weight is assigned to each of the selected rule, in order to combine them into a single estimate of $m(\bx)$. These weights are defined as the ridge regression solution, where each predictor is a rule $\hat{g}_{n, \path}$ for $\smash{\path \in \pathset_{M,n,p_{0}}}$ and weights are constrained to be non-negative. Thus, the aggregated estimate $\hat{m}_{M,n,p_{0}}(\bx)$ of $m(\bx)$ computed in the \textbf{fourth step} of SIRUS has the form
\begin{align} \label{eq_eta}
\hat{m}_{M,n,p_{0}}(\bx)
= \hat{\beta}_0 + \sum_{\path \in \pathset_{M,n,p_{0}}} \hat{\beta}_{n, \path} \hat{g}_{n, \path}( \bx),
\end{align}
where $\hat{\beta}_0$ and $\hat{\beta}_{n, \path}$ are the solutions of the ridge regression problem. More precisely, denoting by $\smash{\boldsymbol{\hat{\beta}}_{n,p_0}}$ the column vector whose components are the coefficients $\smash{\hat{\beta}_{n, \path}}$ for $\smash{\path \in \pathset_{M,n,p_{0}}}$, and letting $\mathbf{Y} = (Y_1,\hdots,Y_n)^T$ and $\boldsymbol{\Gamma}_{n,p_0}$ the matrix whose rows are the rule values $\hat{g}_{n, \path}( \bX_i)$ for $i \in \{1, \hdots,n\}$, we have
\begin{align*}
(\boldsymbol{\hat{\beta}}_{n,p_0}, \hat{\beta}_0) = \underset{\boldsymbol{\beta} \geq 0,\beta_0}{\textrm{argmin}} \frac{1}{n}&||\mathbf{Y} - \beta_0 \boldsymbol{1_n} - \boldsymbol{\Gamma}_{n,p_0}\boldsymbol{\beta}||_2^2 \\[-0.7em] &+ \lambda||\boldsymbol{\beta}||_2^2,
\end{align*}
where $\boldsymbol{1_n} = (1,\hdots,1)^T$ is the $n$-vector with all components equal to $1$, and $\lambda$ is a positive parameter tuned by cross-validation that controls the penalization severity. 
The mininum is taken over $\beta_0 \in \R$ and all the vectors $\smash{\boldsymbol{\beta} = \{\beta_1,\hdots,\beta_{c_n}\} \in \R_+^{c_n}}$ where $\smash{c_n=|\pathset_{M,n,p_{0}}|}$ is the number of selected rules.
Besides, notice that the rule format with an else clause differs from the standard format in the rule learning literature. This modification provides good properties of stability and modularity (investigation of the rules one by one \citep{murdoch2019interpretable}) to SIRUS---see Section $5$ of the Supplementary Material.

This linear rule aggregation is a critical step and deserves additional comments.
Indeed, in RuleFit, the rules are also extracted from a tree ensemble, but aggregated using the Lasso. However, the extracted rules are strongly correlated by construction, and the Lasso selection is known to be highly unstable in such correlated setting. This is the main reason of the instability of RuleFit, as the experiments will show. On the other hand, the sparsity of SIRUS is controlled by the parameter $p_0$, and the ridge regression enables a stable aggregation of the rules.
Furthermore, the constraint $\boldsymbol{\beta} \geq 0$ is added to ensure that all coefficients are non-negative, as in Node harvest \citep{meinshausen2010node}.
Also because of the rule correlation, an unconstrained regression would lead to negative values for some of the coefficients $\smash{\hat{\beta}_{n, \path}}$, and such behavior drastically undermines the interpretability of the algorithm.

\paragraph{Interpretability.}
As stated in the introduction, despite the lack of a precise definition of interpretable models, there are three minimum requirements to be taken into account: simplicity, stability, and predictivity.
These notions need to be formally defined and quantified to enable comparison between algorithms.
\textbf{Simplicity} refers to the model complexity, in particular the number of operations involved in the prediction mechanism. In the case of rule algorithms, a measure of simplicity is naturally given by the number of rules.
Intuitively, a rule algorithm is \textbf{stable} when two independent estimations based on two independent samples return similar lists of rules. Formally, let $\smash{\pathset_{M,n,p_{0}}'}$ be the list of rules output by SIRUS fit on an independent sample $\Dn'$. Then the proportion of rules shared by $\smash{\pathset_{M,n,p_{0}}}$ and $\smash{\pathset_{M,n,p_{0}}'}$ gives a stability measure. Such a metric is known as the Dice-Sorensen index, and is often used to assess variable selection procedures \citep{chao2006abundance,zucknick2008comparing,boulesteix2009stability,he2010stable,alelyani2011dilemma}. In our case, the Dice-Sorensen index is then defined as
\begin{align*}
\hat{S}_{M,n,p_{0}}=\frac{2\big|\pathset_{M,n,p_{0}}\cap\pathset_{M,n,p_{0}}'\big|}{\big|\pathset_{M,n,p_{0}}\big|+\big|\pathset_{M,n,p_{0}}'\big|}.
\end{align*}
However, in practice one rarely has access to an additional sample $\Dn'$. Therefore, to circumvent this problem, we use a $10$-fold cross-validation to simulate data perturbation.
The stability metric is thus empirically defined as the average proportion of rules shared by two models of two distinct folds of the cross-validation. A stability of $1$ means that the exact same list of rules is selected over the $10$ folds, whereas a stability of $0$ means that all rules are distinct between any $2$ folds. 
For \textbf{predictivity} in regression problems, the proportion of unexplained variance is a natural measure of the prediction error. The estimation is performed by $10$-fold cross-validation.

\section{Experiments} \label{sec_xp}

Experiments are run over $8$ diverse public datasets to demonstrate the improvement of SIRUS over state-of-the-art methods. Table $1$ in Section $6$ of the Supplementary Material provides dataset details.

\paragraph{SIRUS rule set.}
\begin{table*}[t]
    \small
	\begin{center}
		\setlength{\fboxrule}{1.5pt}
		\fbox{\begin{minipage}{0.76\textwidth}
				\textbf{Average } $\textrm{Ozone} = 12$ \hspace*{1.5cm}
				\textbf{Intercept} $= -7.8$ \\[3mm]
				\setlength{\tabcolsep}{2pt}
				\begin{tabular}{c c c c c c c c}
					\textbf{Frequency} \hspace*{5mm} & \multicolumn{6}{c}{\textbf{Rule}} & \hspace*{5mm} \textbf{Weight} \\[1mm]
					0.29 \hspace*{5mm} & \textbf{if } & $\textrm{temp} < 65$ & \textbf{ then } & $\textrm{Ozone} = 7$ & \textbf{ else } & $\textrm{Ozone} = 19$ & \hspace*{5mm} 0.12 \\[0.5mm]
					0.17 \hspace*{5mm} & \textbf{if } & $\textrm{ibt} < 189$ & \textbf{ then } & $\textrm{Ozone} = 7$ & \textbf{ else } & $\textrm{Ozone} = 18$ & \hspace*{5mm} 0.07 \\[0.5mm]
					0.063 \hspace*{5mm} & \textbf{if } & $\left\{\begin{tabular}{c} $\textrm{temp} \geq 65$ \\ \textbf{\&} $\textrm{vis} < 150$ \end{tabular}\right.$ & \textbf{ then } & $\textrm{Ozone} = 20$ & \textbf{ else } &  $\textrm{Ozone} = 7$  & \hspace*{5mm} 0.31 \\[0.5mm]	
					0.061 \hspace*{5mm} & \textbf{if } & $\textrm{vh} < 5840$ & \textbf{ then } & $\textrm{Ozone} = 10$ & \textbf{ else } & $\textrm{Ozone} = 20$ & \hspace*{5mm} 0.072 \\[0.5mm]
					0.060 \hspace*{5mm} & \textbf{if } & $\textrm{ibh} < 2110$ & \textbf{ then } & $\textrm{Ozone} = 16$ & \textbf{ else } & $\textrm{Ozone} = 7$ & \hspace*{5mm} 0.14 \\[0.5mm]
					0.058 \hspace*{5mm} & \textbf{if } & $\textrm{ibh} < 2960$ & \textbf{ then } & $\textrm{Ozone} = 15$ & \textbf{ else } & $\textrm{Ozone} = 6$ & \hspace*{5mm} 0.10 \\[0.5mm]
					0.051 \hspace*{5mm} & \textbf{if } & $\left\{\begin{tabular}{c} $\textrm{temp} \geq 65$ \\ \textbf{\&} $\textrm{ibh} < 2110$ \end{tabular}\right.$ & \textbf{ then }&  $\textrm{Ozone} = 21$ & \textbf{ else } & $\textrm{Ozone} = 8$  & \hspace*{5mm} 0.16 \\[0.5mm]	
					0.048 \hspace*{5mm} & \textbf{if } & $\textrm{vis} < 150$ & \textbf{ then } & $\textrm{Ozone} = 14$ & \textbf{ else } & $\textrm{Ozone} = 7$ & \hspace*{5mm} 0.18 \\[0.5mm]
					0.043 \hspace*{5mm} & \textbf{if } & $\left\{\begin{tabular}{c} $\textrm{temp} < 65$ \\ \textbf{\&} $\textrm{ibt} < 120$ \end{tabular}\right.$ & \textbf{ then } & $\textrm{Ozone} = 5$ & \textbf{ else } & $\textrm{Ozone} = 15$ & \hspace*{5mm} 0.15 \\[0.5mm]			
					0.040 \hspace*{5mm} & \textbf{if } & $\textrm{temp} < 70$ & \textbf{ then } & $\textrm{Ozone} = 8$ & \textbf{ else } & $\textrm{Ozone} = 20$ & \hspace*{5mm} 0.14 \\[0.5mm]
					0.039 \hspace*{5mm} & \textbf{if } & $\textrm{ibt} < 227$ & \textbf{ then } & $\textrm{Ozone} = 9$ & \textbf{ else } & $\textrm{Ozone} = 22$ & \hspace*{5mm} 0.21 \\	
				\end{tabular}
		\end{minipage}}
	\end{center}
	\caption{\small{SIRUS rule list for the ``LA Ozone'' dataset (about $9000$ trees are grown to reach convergence).}}
	\label{fig_model}
\end{table*}
Our algorithm is illustrated on the ``LA Ozone'' dataset from \citet{friedman2001elements}, which records the level of atmospheric ozone concentration from eight daily meteorological measurements made in Los Angeles in 1976: wind speed (``wind''), humidity (``humidity''), temperature (``temp''),  inversion base height (``ibh''), daggot pressure gradient (``dpg''), inversion base temperature (``ibt''), visibility (``vis''), and day of the year (``doy''). The response ``Ozone'' is the log of the daily maximum of ozone concentration. 
The list of rules output for this dataset is presented in Table \ref{fig_model}. The column ``Frequency'' refers to $\hat{p}_{M,n}(\path)$, the occurrence frequency of each rule in the forest, used for rule selection.
It enables to grasp how weather conditions impact the ozone concentration. In particular, a temperature larger than $65$\textdegree F or a high inversion base temperature result in high ozone concentrations. The third rule tells us that the interaction of a high temperature with a visibility lower than $150$ miles generates even higher levels of ozone concentration. Interestingly, according to the ninth rule, especially low ozone concentrations are reached when a low temperature and and a low inversion base temperature are combined.
Recall that to generate a prediction for a given query point $\bx$, for each rule the corresponding ozone concentration is retrieved depending on whether $\bx$ satisfies the rule conditions. Then all rule outputs for $\bx$ are multiplied by their associated weight and added together. One can observe that rule importances and weights are not related. For example, the third rule has a higher weight than the most two important ones. It is clear that rule $3$ has multiple constraints and is therefore more sensitive to data perturbation---hence a smaller frequency of appearance in the forest. On the other hand, its associated variance decrease in CART is more important than for the first two rules, leading to a higher weight in the linear combination. Since rules $5$ and $6$ are strongly correlated, their weights are diluted.

\paragraph{Tuning.}
SIRUS has only one hyperparameter which requires fine tuning: the threshold $p_0$ to control the model size by selecting the most frequent rules in the forest. First, the range of possible values of $p_0$ is set so that the model size varies between $1$ and $25$ rules. This arbitrary upper bound is a safeguard to avoid long and complex list of rules that are difficult to interpret. In practice, this limit of $25$ rules is rarely hit, since the following tuning of $p_0$ naturally leads to compact rule lists.
Thus, $p_0$ is tuned within that range by cross-validation to maximize both stability and predictivity. To find a tradeoff between these two properties, we follow a standard bi-objective optimization procedure as illustrated in Figure \ref{fig_tuning}, and described in Section $2$ of the Supplementary Material: $p_0$ is chosen to be as close as possible to the ideal case of $0$ unexplained variance and $90\%$ stability. This tuning procedure is computationally fast: the cost of about $10$ fits of SIRUS. For a robust estimation of $p_0$, the cross-validation is repeated $10$ times and the median $p_0$ value is selected.
\begin{figure}
	\begin{center}
		\includegraphics[height=6cm,width=6cm]{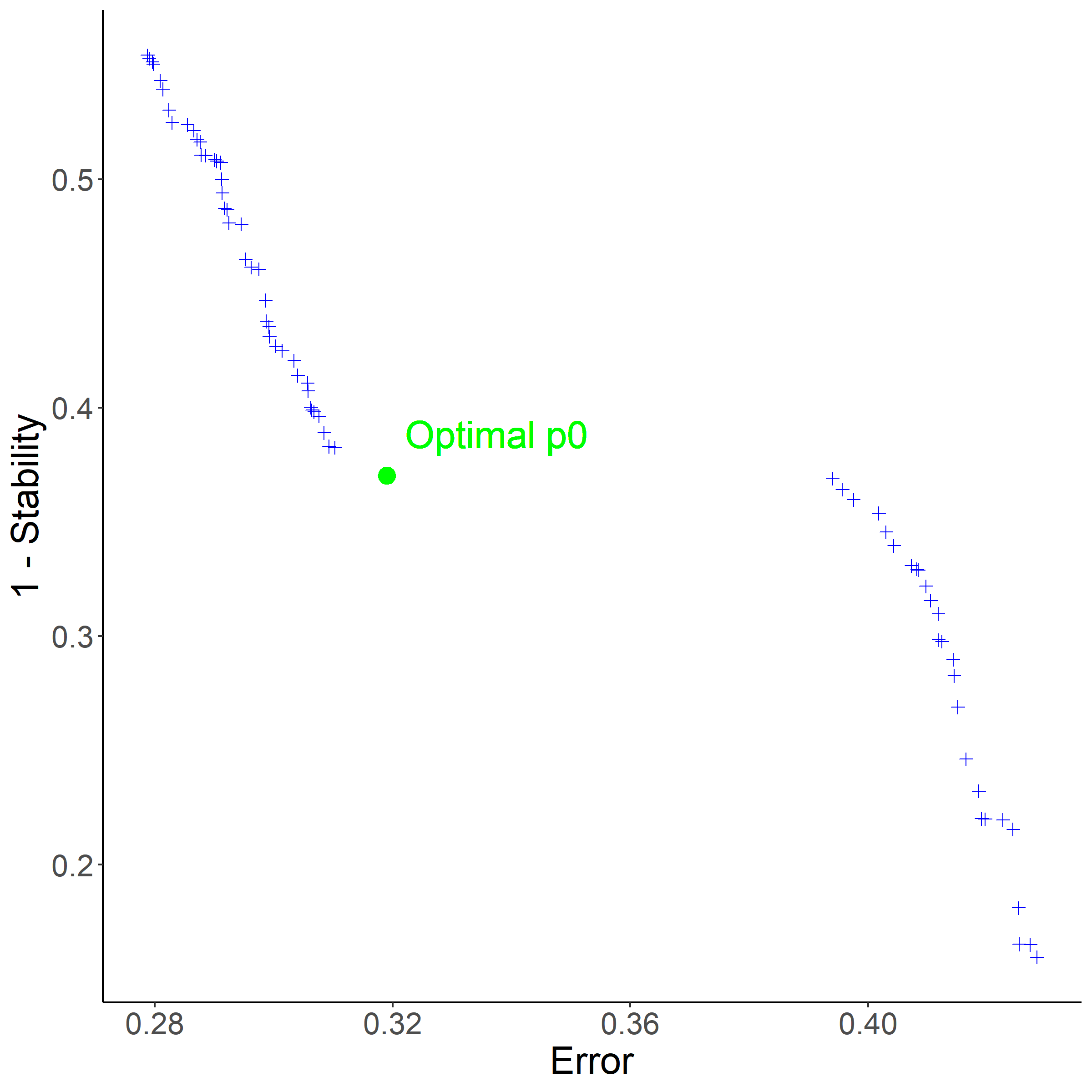}
		\caption{Pareto front of stability versus error when $p_0$ varies for the ``Ozone'' dataset (optimal value in green).}
		\label{fig_tuning}
		\vspace*{-0.7cm}
	\end{center}
\end{figure}
Besides, the optimal number of trees $M$ is set automatically by SIRUS: as stability, predictivity, and computation time increase with the number of trees, no fine tuning is required for $M$. Thus, a stopping criterion is designed to grow the minimum number of trees which enforces that stability and predictivity are greater than $95\%$ of their maximum values (reached when $M \rightarrow \infty$)---see Section $7$ of the Supplementary Material for a detailed definition of this criterion.
Finally, we use the standard settings of random forests (well-known for their excellent performance, in particular \textit{mtry} is $\lfloor p/3 \rfloor$ and at least $2$), and set $q = 10$ quantiles, while categorical variables are handled as natively defined in trees.

\paragraph{Performance.}
We compare SIRUS with its two main competitors RuleFit (with rule predictors only) and Node harvest.
For predictive accuracy, we ran random forests and (pruned) CART to provide the baseline. Only to compute stability metrics, data is binned using $10$ quantiles to fit Rulefit and Node harvest.
Our \texttt{R}/\texttt{C++} package \texttt{sirus} (available from \texttt{CRAN}) is adapted from \texttt{ranger}, a fast random forests implementation \citep{wright2017ranger}. We also use available \texttt{R} implementations \texttt{pre} \citep[RuleFit]{fokkema2017pre} and \texttt{nodeharvest} \citep{meinshausen2015package}.
While the predictive accuracy of SIRUS is comparable to Node harvest and slightly below RuleFit, the stability is considerably improved with much smaller rule lists.
Experimental results are gathered in Table \ref{table_stability_2obj}a for model sizes, Table \ref{table_stability_2obj}b for stability, and Table \ref{table_error_2obj} for predictive accuracy.
All results are averaged over $10$ repetitions of the cross-validation procedure. Since standard deviations are negligible, they are not displayed to increase readability.
Besides, in the last column of Table  \ref{table_error_2obj}, $p_0$ is set to increase the number of rules in SIRUS to reach RuleFit and Node harvest model size (about $50$ rules): predictivity is then as good as RuleFit.
Finally, the column ``SIRUS sparse'' of Tables \ref{table_stability_2obj} and \ref{table_error_2obj} shows the excellent behavior of SIRUS in a sparse setting: for each dataset, $3$ randomly permuted copies of each variable are added to the data, leaving SIRUS performance almost untouched.

\begin{figure}
	\begin{center}
		\includegraphics[height=8cm,width=8cm]{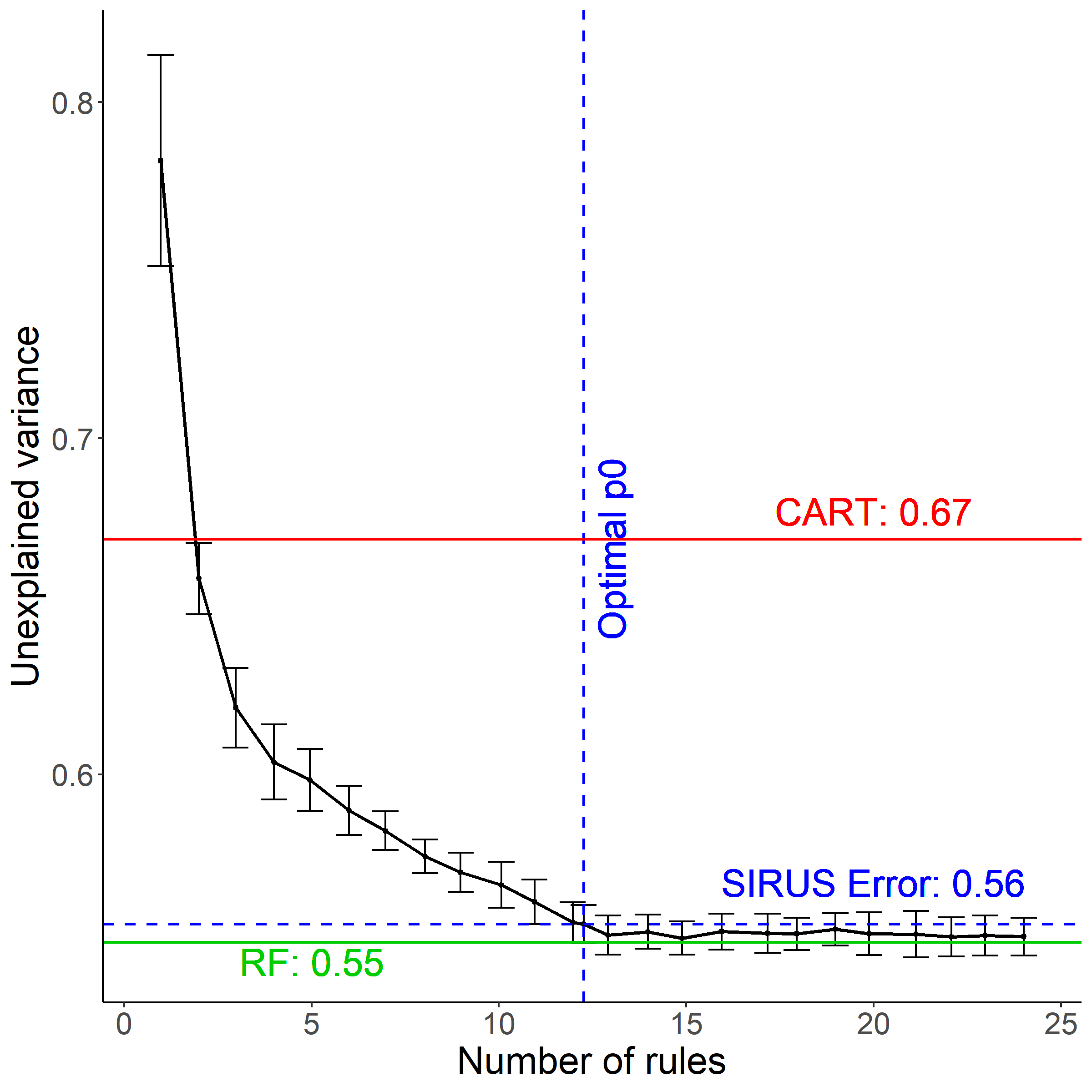}
		\includegraphics[height=8cm,width=8cm]{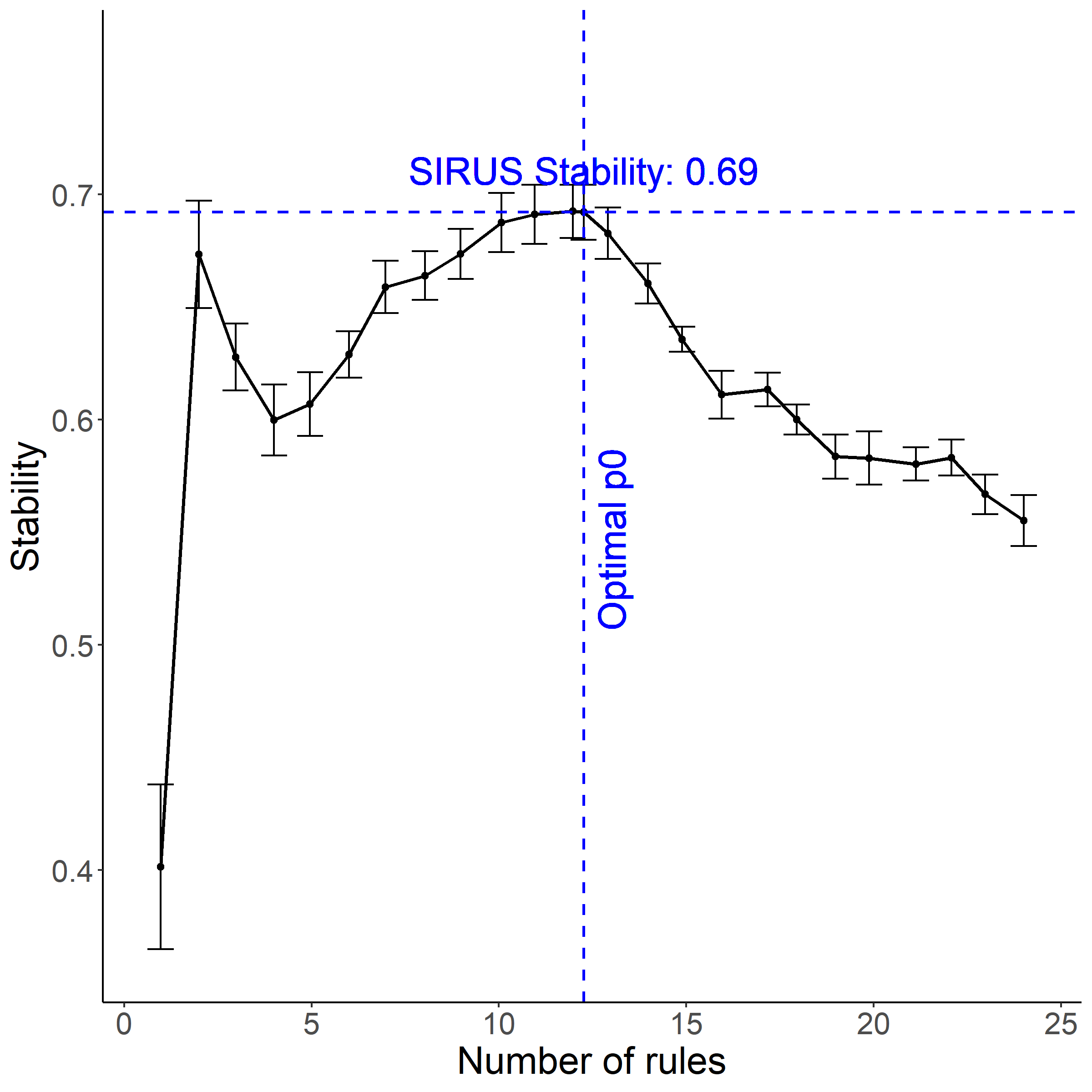}
		\caption{\small{For the dataset ``Diabetes'', unexplained variance (top panel) and stability (bottom panel) versus the number of rules when $p_0$ varies, estimated via 10-fold cross-validation (results are averaged over $10$ repetitions).}}
		\label{fig_diabetes}
	\end{center}
\end{figure}

\begin{figure}
	\begin{center}
		\includegraphics[height=7cm,width=7cm]{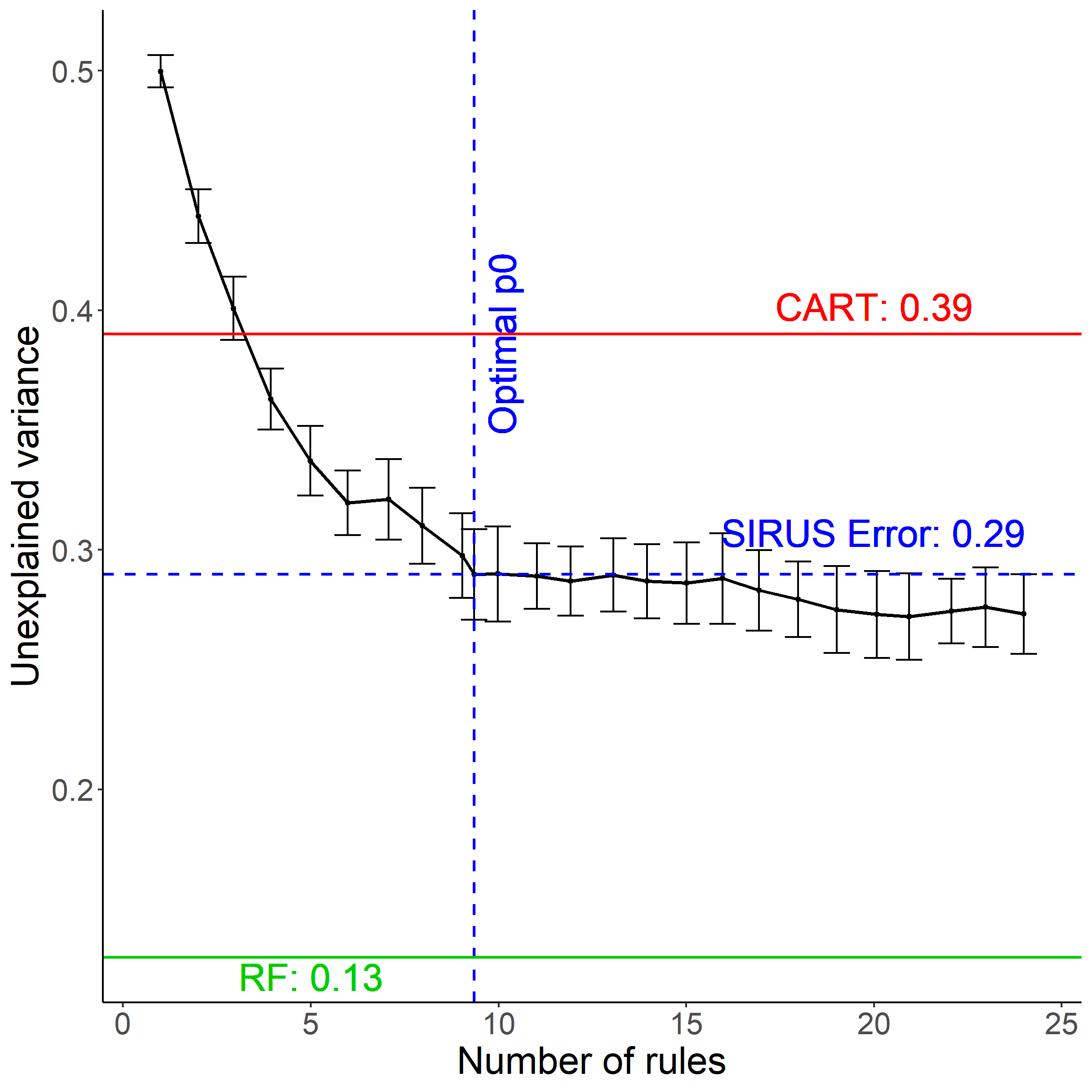}
		\includegraphics[height=7cm,width=7cm]{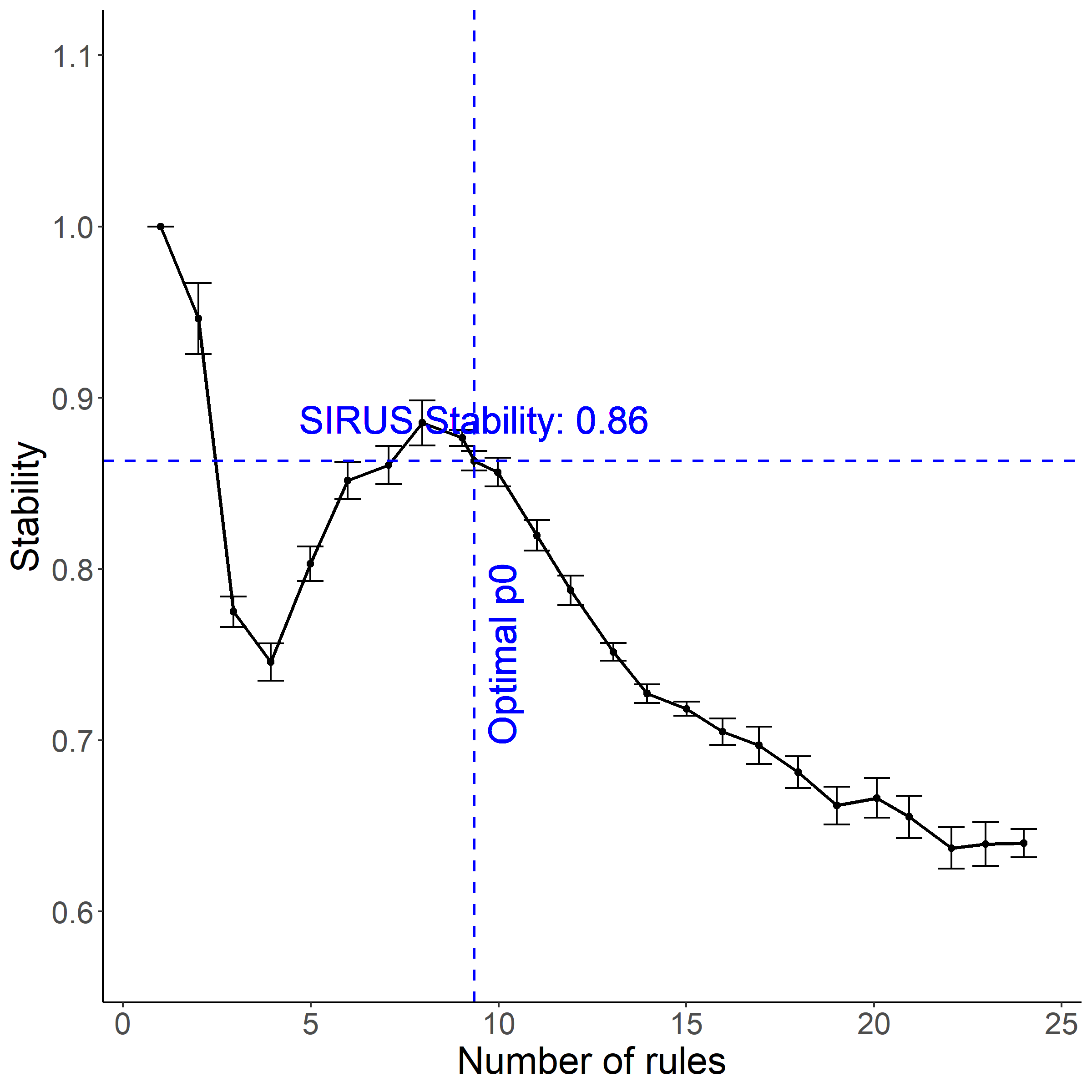}
		\caption{\small{For the dataset ``Machine'', unexplained variance (top panel) and stability (bottom panel) versus the number of rules when $p_0$ varies, estimated via 10-fold cross-validation (results are averaged over $10$ repetitions).}}
		\label{fig_machine}
	\end{center}
\end{figure}

To illustrate the typical behavior of our method, we comment the results for two specific datasets: ``Diabetes'' \citep{efron2004least} and ``Machine'' \citep{Dua:2019}. The ``Diabetes'' data contains $n = 442$ diabetic patients and the response of interest $Y$ is a measure of disease progression over one year. A total of $10$ variables are collected for each patient: age, sex, body mass index, average blood pressure, and six blood serum measurements $s1, s2,\hdots, s6$.
For this dataset, SIRUS is as predictive as a random forest, with only $12$ rules when the forest performs about $10^4$ operations: the unexplained variance is $0.56$ for SIRUS and $0.55$ for random forest. Notice that CART performs considerably worse with $0.67$ unexplained variance.
For the second dataset, ``Machine'', the output $Y$ of interest is the CPU performance of computer hardware. For $n = 209$ machines, $6$ variables are collected about the machine characteristics. For this dataset, SIRUS, RuleFit, and Node harvest have a similar predictivity, in-between CART and random forests. 
Our algorithm achieves such performance with a readable list of only $9$ rules stable at $86\%$, while RuleFit and Node harvest incorporate respectively $44$ and $42$ rules with stability levels of $23\%$ and $29\%$.
Stability and predictivity are represented as $p_0$ varies for ``Diabetes'' and ``Machine'' datasets in Figures \ref{fig_diabetes} and \ref{fig_machine}, respectively. 

\begin{table}
    \setlength{\tabcolsep}{1pt}
	\centering
	\vspace*{-0.4cm}
	\subtable[Model Size]{
    	\begin{tabular}{|c |c | c | c | c | c |}
    		\hline \hline
    		\textbf{Dataset} & \textbf{CART} & \textbf{RuleFit} & \begin{tabular}{c} \textbf{Node} \\ \textbf{harvest}	\end{tabular} & \textbf{SIRUS} & \begin{tabular}{c} \textbf{SIRUS} \\ \textbf{sparse} \end{tabular} \\
    		\hline
    		Ozone & 15 & 21 & 46 & \textbf{11} & 10 \\	
    		Mpg & 15 & 40 & 43 & \textbf{10} & 10 \\	
    		Prostate & 11 & 14 & 41 & \textbf{9} & 12 \\	
    		Housing & 15 & 54 & 40 & \textbf{6} & 6 \\	
    		Diabetes & \textbf{12} & 25 & 42 & \textbf{12} & 15 \\	
    		Machine & \textbf{8} & 44 & 42 & 9 & 7 \\
    		Abalone & 20 & 58 & 35 & \textbf{8} & 13 \\
    		Bones & 17 & 5 & 13 & \textbf{1} & 1 \\
    		\hline \hline
    	\end{tabular}}
	\subtable[Stability]{
    	\begin{tabular}{|c | c | c | c | c | c |}
    		\hline  \hline
    		\textbf{Dataset} & \textbf{RuleFit} & \textbf{Node harvest} & \textbf{SIRUS} & \begin{tabular}{c}
    		     \textbf{SIRUS} \\ \textbf{sparse} \end{tabular}\\
    		\hline
    		Ozone & 0.22 & 0.30 & \textbf{0.62} & 0.63 \\	
    		Mpg & 0.25 & 0.43 & \textbf{0.77} & 0.76 \\	
    		Prostate & 0.32 & 0.23 & \textbf{0.58} & 0.59 \\	
    		Housing  & 0.19 & 0.40 & \textbf{0.82} & 0.82 \\	
    		Diabetes & 0.18 & 0.39 & \textbf{0.69} & 0.65 \\	
    		Machine  & 0.23 & 0.29 & \textbf{0.86} & 0.84 \\	
    		Abalone  & 0.31 & 0.38 & \textbf{0.75} & 0.74 \\	
    		Bones  & 0.59 & 0.52 & \textbf{0.96} & 0.78 \\
    		\hline \hline
    	\end{tabular}}
    	\vspace*{-1mm}
    	\caption{\small{Mean model size and stability over a $10$-fold cross-validation for various public datasets. Minimum size and maximum stability are in bold (``SIRUS sparse'' put aside).}}
    	\label{table_stability_2obj}
\end{table}

\begin{table*}[t]
    \setlength{\tabcolsep}{3pt}
	\centering
	\begin{tabular}{|c | c | c | c | c | c | c | c | c |}
		\hline \hline
		\textbf{Dataset} & \begin{tabular}{c}\textbf{Random} \\ \textbf{Forest} \end{tabular} & \textbf{CART} & \textbf{RuleFit} & \begin{tabular}{c}\textbf{Node} \\ \textbf{harvest}\end{tabular} & \textbf{SIRUS} & \begin{tabular}{c} \textbf{SIRUS} \\ \textbf{sparse} \end{tabular} & \begin{tabular}{c}\textbf{SIRUS} \\ \textbf{50 rules}\end{tabular} \\ 
		\hline
		Ozone & 0.25 & 0.36 & \textbf{0.27} & 0.31 & 0.32 & 0.32 & \textbf{0.26} \\	
		Mpg & 0.13 & 0.20 & \textbf{0.15} & 0.20 & 0.20 & 0.20 & \textbf{0.15} \\		
		Prostate & 0.48 & 0.60 & \textbf{0.53} & \textbf{0.52} & \textbf{0.55} & 0.51 & \textbf{0.54} \\	
		Housing & 0.13 & 0.28 & \textbf{0.16}  & 0.24 & 0.30 & 0.31 & 0.20 \\	
		Diabetes & 0.55 & 0.67 & \textbf{0.55} & \textbf{0.58} & \textbf{0.56} & 0.56 & \textbf{0.55}  \\	
		Machine & 0.13 & 0.39 & \textbf{0.26} & \textbf{0.29} & \textbf{0.29} & 0.32 & \textbf{0.27} \\	
		Abalone & 0.44 & 0.56 & \textbf{0.46} & 0.61 & 0.66 & 0.64 & 0.64 \\	
		Bones & 0.67 & 0.67 & \textbf{0.70} & \textbf{0.70} & \textbf{0.73} & 0.77 & \textbf{0.73}  \\
		\hline \hline
	\end{tabular}
	\vspace*{1mm}
	\caption{\small{Proportion of unexplained variance estimated over a $10$-fold cross-validation for various public datasets. For rule algorithms only, i.e., RuleFit, Node harvest, and SIRUS, minimum values are displayed in bold, as well as values within 10\% of the minimum for each dataset (``SIRUS sparse'' put aside).}}
	\label{table_error_2obj}
\end{table*}


\section{Theoretical Analysis} \label{sec_theory}

Among the three minimum requirements for interpretable models, stability is the critical one. In SIRUS, simplicity is explicitly controlled by the hyperparameter $p_0$. The wide literature on rule learning provides many experiments to show that rule algorithms have an accuracy comparable to tree ensembles. On the other hand, designing a stable rule procedure is more challenging \citep{letham2015interpretable, murdoch2019interpretable}.
For this reason, we therefore focus our theoretical analysis on the asymptotic stability of SIRUS.

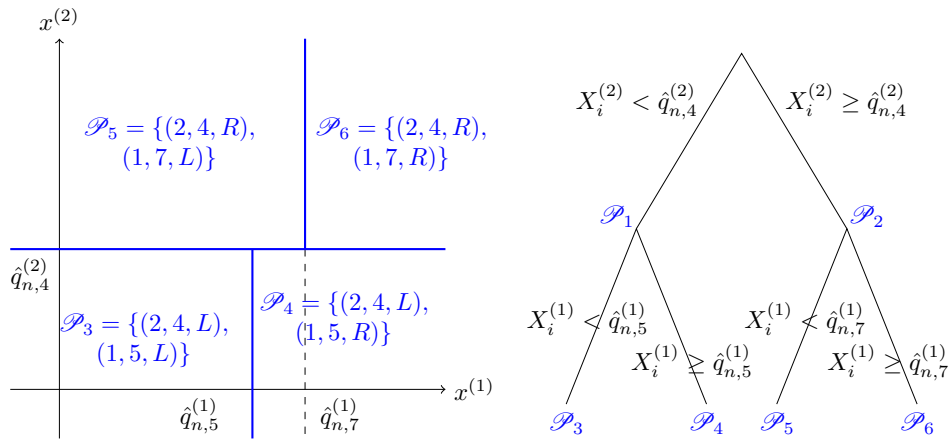
\begin{figure*}
	\centering
	\begin{adjustbox}{width=5in}
		\begin{tikzpicture}
		
		\draw[->] (-0.7,0) -- (5.5,0);
		\draw (5.5,0) node[right] {$x^{(1)}$};
		\draw [->] (0,-0.7) -- (0,5);
		\draw (0,5) node[above] {$x^{(2)}$};
		\draw [color = blue, line width = 0.3 mm] (3.5,2) -- (3.5,5);
		\draw [dashed] (3.5,2) -- (3.5,-0.7);
		\draw [color = blue, line width = 0.3 mm] (2.75,2) -- (2.75,-0.7);
		\draw (4,0) node[below] {$\hat{q}_{n,7}^{(1)}$};
		\draw (2,0) node[below] {$\hat{q}_{n,5}^{(1)}$};
		\draw [color = blue, line width = 0.3 mm] (5.5,2) -- (-0.7,2);
		\draw (0,2) node[below left] {$\hat{q}_{n,4}^{(2)}$};
		\draw (0,3.5) node[right, color = blue, text width=2cm] {\begin{tabular}{c} $\path_5 =\{ (2,4,R),$ \\ $(1,7,L)\}$ \end{tabular}};
		\draw (3.3,3.5) node[right, color = blue] {\begin{tabular}{c} $\path_6 =\{ (2,4,R),$ \\ $(1,7,R)\}$ \end{tabular}};
		\draw (-0.35,0.7) node[right, color = blue] {\begin{tabular}{c} $\path_3 =\{ (2,4,L),$ \\ $(1,5,L)\}$ \end{tabular}};
		\draw (2.5,1) node[right, color = blue] {\begin{tabular}{c} $\path_4 =\{ (2,4,L),$ \\ $(1,5,R)\}$ \end{tabular}};
		\end{tikzpicture}
		
		\begin{tikzpicture}
		\draw (7.5,2.5) --(9,5) --(10.5,2.5);
		\node[below left] at (8.5,4.7) {$X_i^{(2)} < \hat{q}_{n,4}^{(2)}$};
		\node[below right] at (9.5,4.7) {$X_i^{(2)} \geq \hat{q}_{n,4}^{(2)}$};
		\node[above left, color = blue] at (7.6,2.4) {$\path_1$};
		\node[above right, color = blue] at (10.4,2.4) {$\path_2$};
		\draw (9.5,0) --(10.5,2.5) --(11.5,0);
		\node[left] at (10.9,1.2) {$X_i^{(1)} < \hat{q}_{n,7}^{(1)}$};
		\node[right] at (10.1,0.6) {$X_i^{(1)} \geq \hat{q}_{n,7}^{(1)}$};
		\node[below, color = blue] at (9.5,0) {$\path_5$};
		\node[below, color = blue] at (11.5,0) {$\path_6$};
		\draw (6.5,0) --(7.5,2.5) --(8.5,0);
		\node[left] at (7.8,1.2) {$X_i^{(1)} < \hat{q}_{n,5}^{(1)}$};
		\node[right] at (7.3,0.6) {$X_i^{(1)} \geq \hat{q}_{n,5}^{(1)}$};
		\node[below, color = blue] at (6.5,0) {$\path_3$};
		\node[below, color = blue] at (8.5,0) {$\path_4$};
		
		\end{tikzpicture}
	\end{adjustbox}
	\caption{\small{Example of a root node $\R^2$ partitioned by a randomized tree of depth 2: the tree on the right, the associated paths and hyperrectangles of length $d=2$ on the left.}}
	\label{fig_paths}
	\vspace*{-0.5cm}
\end{figure*}

To get started, we need a rigorous definition of the rule extraction procedure. To this aim, we introduce a symbolic representation of a path in a tree, which describes the sequence of splits to reach a given (inner or terminal) node from the root. We insist that such path encoding can be used in both the empirical and theoretical algorithms to define rules. A path $\path$ is defined as
\[
\mathscr{P} =\{ (j_{k},r_{k},s_{k}), \, k = 1,\hdots,d\},
\]
where $d$ is the tree depth, and for $k \in \{1,\hdots,d\}$, the triplet $(j_{k},r_{k},s_{k})$ describes how to move from level $(k-1)$ to level $k$, with a split using the coordinate $j_k\in \{1,\hdots,p\}$, the index $r_{k}\in \{1,\hdots,q - 1\}$ of the corresponding quantile, and
a side $s_{k} = L$ if we go to the left and $s_{k} = R$ if we go to the right---see Figure \ref{fig_paths}. The set of all possible such paths is denoted by $\Pi$.
Each tree of the forest is randomized in two ways: $(i)$ the sample $\Dn$ is bootstrapped prior to the construction of the tree, and $(ii)$ a subset of coordinates is randomly selected to find the best split at each node. This randomization mechanism is governed by a random variable that we call $\Theta$. We define $T(\Theta,\mathscr {D}_{n})$, a random subset of $\Pi$, as the collection of the extracted paths from the random tree built with $\Theta$ and $\mathscr {D}_{n}$.
Now, let $\Theta_1, \hdots, \Theta_{\ell}, \hdots, \Theta_M$ be the independent randomizations of the $M$ trees of the forest.
With this notation, the empirical frequency of occurrence of a path $\path \in \Pi$ in the forest takes the form
\[
\hat{p}_{M,n}(\path)=\frac{1}{M}\sum_{\ell=1}^{M}\mathds{1}_{\path\in T(\Theta_{\ell},\mathscr{D}_{n})},
\]
which is simply the proportion of trees that contain $\path$. By definition, $\hat{p}_{M,n}(\path)$ is the Monte Carlo estimate of the probability $p_n(\mathscr{P})$ that a $\Theta$-random tree contains a particular path $\mathscr{P} \in \Pi$, that is,
\[
p_n(\mathscr{P})=\mathbb{P}(\mathscr{P}\in T(\Theta,\mathscr{D}_{n})|\mathscr{D}_{n}).
\]

Next, we introduce all theoretical counterparts of the empirical quantities involved in SIRUS, which do not depend on the sample $\mathscr{D}_{n}$ but only on the unknown distribution of $(\bX,Y)$.
We let $T^{\star}(\Theta)$ be the list of all paths contained in the theoretical tree built with randomness $\Theta$, in which splits are chosen to maximize the theoretical CART-splitting criterion instead of the empirical one.
The probability $p^{\star}(\path)$ that a given path $\path$ belongs to a theoretical randomized tree (the theoretical counterpart of $p_n(\path)$) is
\[
p^{\star}(\path)=\mathds{P}(\path\in T^{\star}(\Theta)).
\]
We finally define the theoretical set of selected paths  
$\mathscr{P}^{\star}_{p_0}=\{ \path \in \Pi : p^{\star}(\path) > p_0\}$ (with the same post-treatment as for the data-based procedure---see Section \ref{sec_sirus}---to remove linear dependence between rules, and discarding paths with a null coefficient in the rule aggregation).
As it is often the case in the theoretical analysis of random forests, \citep{scornet2015consistency, mentchquantifying2016}, we assume throughout this section that the subsampling of $a_n$ observations prior to each tree construction is done without replacement to alleviate the mathematical analysis. Our stability result holds under the following mild assumptions:
\begin{enumerate}
	\item[(A1)] The subsampling rate $a_n$ satisfies $\lim\limits_{n \to \infty} a_{n}=\infty$ and $\lim\limits_{n \to \infty} \frac{a_{n}}{n}=0$, and the number of trees $M_n$ satisfies $\smash{\lim\limits_{n \to \infty} M_n = \infty}$.
	\item[(A2)] The random variable $\bX$ has a strictly positive density $f$ with respect to the Lebesgue measure on $\mathds R^p$. Furthermore, for all $j\in \{1,\hdots,p\}$, the marginal density $f^{(j)}$ of $X^{(j)}$ is continuous, bounded, and strictly positive. Finally, the random variable $Y$ is bounded.
\end{enumerate}

\begin{theorem}
	\label{theorem_stability}
	Assume that Assumptions (A1) and (A2) are satisfied, and let $\setpstar = \{ p^{\star}(\path): \path \in \Pi \}$ be the set of all theoretical probabilities of appearance for each path $\path$. Then, provided $p_{0} \in [0, 1] \setminus \setpstar$ and $\lambda > 0$, we have 
	\[
	\lim \limits_{n \to \infty}  \hat{S}_{M_{n},n,p_{0}} = 1 \quad \mbox{in probability}.
	\]
\end{theorem}
Theorem \ref{theorem_stability} states that SIRUS is stable: provided that the sample size is large enough, the same list of rules is systematically output across several fits on independent samples. The analysis conducted in the proof---Section $1$ of the Supplementary Material---highlights that the cut discretization (performed at quantile values only), as well as considering random forests (instead of boosted tree ensembles as in RuleFit) are the cornerstones to stabilize rule models extracted from tree ensembles. Furthermore, the experiments in Section \ref{sec_xp} show the high empirical stability of SIRUS in finite-sample regimes.

\section{Conclusion}
\label{conclusion}
Interpretability of machine learning algorithms is required whenever the targeted applications involve critical decisions. Although interpretability does not have a precise definition, we argued that simplicity, stability, and predictivity are minimum requirements for interpretable models. In this context, rule algorithms are well known for their good predictivity and simple structures, but also to be often highly unstable.
Therefore, we proposed a new regression rule algorithm called SIRUS, whose general principle is to extract rules from random forests. Our algorithm exhibits an accuracy comparable to state-of-the-art rule algorithms, while producing much more stable and shorter lists of rules. This remarkably stable behavior is theoretically understood since the rule selection is consistent. A \texttt{R/C++} software \texttt{sirus} is available from \texttt{CRAN}.

\section*{Acknowledgements}
We thank the reviewers for their insightful comments and suggestions.

\bibliography{biblio}

\begin{thebibliography}{}

\bibitem[Alelyani et~al., 2011]{alelyani2011dilemma}
Alelyani, S., Zhao, Z., and Liu, H. (2011).
\newblock A dilemma in assessing stability of feature selection algorithms.
\newblock In {\em 13th IEEE International Conference on High Performance
  Computing \& Communication}, pages 701--707, Piscataway. IEEE.

\bibitem[B{\'e}nard et~al., 2021]{benard2020sirus}
B{\'e}nard, C., Biau, G., Da~Veiga, S., and Scornet, E. (2021).
\newblock Sirus: Stable and interpretable rule set for classification.
\newblock {\em Electronic Journal of Statistics}, 15:427--505.

\bibitem[Boulesteix and Slawski, 2009]{boulesteix2009stability}
Boulesteix, A.-L. and Slawski, M. (2009).
\newblock Stability and aggregation of ranked gene lists.
\newblock {\em Briefings in Bioinformatics}, 10:556--568.

\bibitem[Breiman, 2001a]{breiman2001random}
Breiman, L. (2001a).
\newblock Random forests.
\newblock {\em Machine Learning}, 45:5--32.

\bibitem[Breiman, 2001b]{breiman2001statistical}
Breiman, L. (2001b).
\newblock Statistical modeling: The two cultures (with comments and a rejoinder
  by the author).
\newblock {\em Statistical Science}, 16:199--231.

\bibitem[Breiman et~al., 1984]{leo1984classification}
Breiman, L., Friedman, J., Olshen, R., and Stone, C. (1984).
\newblock {\em Classification and Regression Trees}.
\newblock Chapman \& Hall/CRC, Boca Raton.

\bibitem[Chao et~al., 2006]{chao2006abundance}
Chao, A., Chazdon, R., Colwell, R., and Shen, T.-J. (2006).
\newblock Abundance-based similarity indices and their estimation when there
  are unseen species in samples.
\newblock {\em Biometrics}, 62:361--371.

\bibitem[Clark and Niblett, 1989]{clark1989cn2}
Clark, P. and Niblett, T. (1989).
\newblock The {CN2} induction algorithm.
\newblock {\em Machine Learning}, 3:261--283.

\bibitem[Cohen, 1995]{cohen1995fast}
Cohen, W. (1995).
\newblock Fast effective rule induction.
\newblock In {\em Proceedings of the 12th International Conference on Machine
  Learning}, pages 115--123, San Francisco. Morgan Kaufmann Publishers Inc.

\bibitem[Cohen and Singer, 1999]{cohen1999simple}
Cohen, W. and Singer, Y. (1999).
\newblock A simple, fast, and effective rule learner.
\newblock In {\em Proceedings of the 16th National Conference on Artificial
  Intelligence and 11th Conference on Innovative Applications of Artificial
  Intelligence}, pages 335--342, Palo Alto. AAAI Press.

\bibitem[Dembczy{\'n}ski et~al., 2010]{dembczynski2010ender}
Dembczy{\'n}ski, K., Kot{\l}owski, W., and S{\l}owi{\'n}ski, R. (2010).
\newblock {ENDER}: A statistical framework for boosting decision rules.
\newblock {\em Data Mining and Knowledge Discovery}, 21:52--90.

\bibitem[Dua and Graff, 2017]{Dua:2019}
Dua, D. and Graff, C. (2017).
\newblock {UCI} machine learning repository.

\bibitem[Efron et~al., 2004]{efron2004least}
Efron, B., Hastie, T., Johnstone, I., and Tibshirani, R. (2004).
\newblock Least angle regression.
\newblock {\em The Annals of statistics}, 32:407--499.

\bibitem[Fokkema, 2017]{fokkema2017pre}
Fokkema, M. (2017).
\newblock {PRE}: An {R} package for fitting prediction rule ensembles.
\newblock {\em arXiv:1707.07149}.

\bibitem[Frank and Witten, 1998]{frank1998generating}
Frank, E. and Witten, I.~H. (1998).
\newblock Generating accurate rule sets without global optimization.
\newblock In {\em Proceedings of the 15th International Conference on Machine
  Learning}, pages 144--151, San Francisco. Morgan Kaufmann Publishers Inc.

\bibitem[Friedman et~al., 2001]{friedman2001elements}
Friedman, J., Hastie, T., and Tibshirani, R. (2001).
\newblock {\em The Elements of Statistical Learning}, volume~1.
\newblock Springer, New York.

\bibitem[Friedman et~al., 2010]{friedman2010regularization}
Friedman, J., Hastie, T., and Tibshirani, R. (2010).
\newblock Regularization paths for generalized linear models via coordinate
  descent.
\newblock {\em Journal of statistical software}, 33:1.

\bibitem[Friedman and Popescu, 2003]{friedman2003importance}
Friedman, J. and Popescu, B. (2003).
\newblock Importance sampled learning ensembles.
\newblock {\em Journal of Machine Learning Research}, 94305:1--32.

\bibitem[Friedman and Popescu, 2008]{friedman2008predictive}
Friedman, J. and Popescu, B. (2008).
\newblock Predictive learning via rule ensembles.
\newblock {\em The Annals of Applied Statistics}, 2:916--954.

\bibitem[F{\"u}rnkranz and Widmer, 1994]{furnkranz1994incremental}
F{\"u}rnkranz, J. and Widmer, G. (1994).
\newblock Incremental reduced error pruning.
\newblock In {\em Proceedings of the 11th International Conference on Machine
  Learning}, pages 70--77, San Francisco. Morgan Kaufmann Publishers Inc.

\bibitem[He and Yu, 2010]{he2010stable}
He, Z. and Yu, W. (2010).
\newblock Stable feature selection for biomarker discovery.
\newblock {\em Computational Biology and Chemistry}, 34:215--225.

\bibitem[Letham et~al., 2015]{letham2015interpretable}
Letham, B., Rudin, C., McCormick, T., and Madigan, D. (2015).
\newblock Interpretable classifiers using rules and {B}ayesian analysis:
  Building a better stroke prediction model.
\newblock {\em The Annals of Applied Statistics}, 9:1350--1371.

\bibitem[Lipton, 2016]{lipton2016mythos}
Lipton, Z. (2016).
\newblock The mythos of model interpretability.
\newblock {\em arXiv:1606.03490}.

\bibitem[Louppe, 2014]{louppe2014understanding}
Louppe, G. (2014).
\newblock Understanding random forests: From theory to practice.
\newblock {\em arXiv preprint arXiv:1407.7502}.

\bibitem[Margot et~al., 2018]{margot2018rule}
Margot, V., Baudry, J.-P., Guilloux, F., and Wintenberger, O. (2018).
\newblock Rule induction partitioning estimator.
\newblock In {\em Proceedings of the 14th International Conference on Machine
  Learning and Data Mining in Pattern Recognition}, pages 288--301, New York.
  Springer.

\bibitem[Margot et~al., 2019]{margot2019consistent}
Margot, V., Baudry, J.-P., Guilloux, F., and Wintenberger, O. (2019).
\newblock Consistent regression using data-dependent coverings.
\newblock {\em arXiv:1907.02306}.

\bibitem[Meinshausen, 2010]{meinshausen2010node}
Meinshausen, N. (2010).
\newblock Node harvest.
\newblock {\em The Annals of Applied Statistics}, 4:2049--2072.

\bibitem[Meinshausen, 2015]{meinshausen2015package}
Meinshausen, N. (2015).
\newblock Package ‘nodeharvest’.

\bibitem[Mentch and Hooker, 2016]{mentchquantifying2016}
Mentch, L. and Hooker, G. (2016).
\newblock Quantifying uncertainty in random forests via confidence intervals
  and hypothesis tests.
\newblock {\em Journal of Machine Learning Research}, 17:841--881.

\bibitem[Murdoch et~al., 2019]{murdoch2019interpretable}
Murdoch, W., Singh, C., Kumbier, K., Abbasi-Asl, R., and Yu, B. (2019).
\newblock Interpretable machine learning: Definitions, methods, and
  applications.
\newblock {\em arXiv:1901.04592}.

\bibitem[Quinlan, 1992]{quinlan1992c4}
Quinlan, J. (1992).
\newblock {\em C4.5: Programs for Machine Learning}.
\newblock Morgan Kaufmann, San Mateo.

\bibitem[Rivest, 1987]{rivest1987learning}
Rivest, R. (1987).
\newblock Learning decision lists.
\newblock {\em Machine Learning}, 2:229--246.

\bibitem[Scornet et~al., 2015]{scornet2015consistency}
Scornet, E., Biau, G., and Vert, J.-P. (2015).
\newblock Consistency of random forests.
\newblock {\em The Annals of Statistics}, 43(4):1716--1741.

\bibitem[Tibshirani, 1996]{tibshirani1996regression}
Tibshirani, R. (1996).
\newblock Regression shrinkage and selection via the lasso.
\newblock {\em Journal of the Royal Statistical Society. Series B}, pages
  267--288.

\bibitem[Van~der Vaart, 2000]{van2000asymptotic}
Van~der Vaart, A. (2000).
\newblock {\em Asymptotic Statistics}, volume~3.
\newblock Cambridge University Press, Cambridge.

\bibitem[Wei et~al., 2019]{wei2019generalized}
Wei, D., Dash, S., Gao, T., and G{\"u}nl{\"u}k, O. (2019).
\newblock Generalized linear rule models.
\newblock {\em arXiv preprint arXiv:1906.01761}.

\bibitem[Weiss and Indurkhya, 2000]{weiss2000lightweight}
Weiss, S. and Indurkhya, N. (2000).
\newblock Lightweight rule induction.
\newblock In {\em Proceedings of the 17th International Conference on Machine
  Learning}, pages 1135--1142, San Francisco. Morgan Kaufmann Publishers Inc.

\bibitem[Wright and Ziegler, 2017]{wright2017ranger}
Wright, M. and Ziegler, A. (2017).
\newblock ranger: A fast implementation of random forests for high dimensional
  data in {C}++ and {R}.
\newblock {\em Journal of Statistical Software}, 77:1--17.

\bibitem[Yu, 2013]{yu2013stability}
Yu, B. (2013).
\newblock Stability.
\newblock {\em Bernoulli}, 19:1484--1500.

\bibitem[Yu and Kumbier, 2019]{yu2019three}
Yu, B. and Kumbier, K. (2019).
\newblock Three principles of data science: Predictability, computability, and
  stability ({PCS}).
\newblock {\em arXiv:1901.08152}.

\bibitem[Zucknick et~al., 2008]{zucknick2008comparing}
Zucknick, M., Richardson, S., and Stronach, E. (2008).
\newblock Comparing the characteristics of gene expression profiles derived by
  univariate and multivariate classification methods.
\newblock {\em Statistical Applications in Genetics and Molecular Biology},
  7:1--34.

\end{thebibliography}

\onecolumn
\title{\textbf{Supplementary Material For: Interpretable Random Forests via Rule Extraction}}
\author{}
\date{}

\maketitle

\setcounter{section}{0}

\section{Proof of Theorem 1: Asymptotic Stability} \label{appendix_proof}

\begin{proof}[Proof of Theorem $1$]

    We recall that stability is assessed by the Dice-Sorensen index as
    \begin{align*}
    \hat{S}_{M,n,p_{0}}=\frac{2\big|\pathset_{M,n,p_{0}}\cap\pathset_{M,n,p_{0}}'\big|}{\big|\pathset_{M,n,p_{0}}\big|+\big|\pathset_{M,n,p_{0}}'\big|}, 
    \end{align*}
    where $\pathset_{M,n,p_{0}}'$ stands for the list of rules output by SIRUS fit with an independent sample $\Dn'$ and where the random forest is parameterized by independent copies $\smash{\Theta_1',\hdots,\Theta_M'}$. 
    
	We consider $p_0 \in [0, 1] \setminus \setpstar$ and $\lambda > 0$.
	There are two sources of randomness in the estimation of the final set of selected paths: $(i)$ the path extraction from the random forest based on $\smash{\hat{p}_{M,n}(\path)}$ for $\path \in \Pi$, and $(ii)$ the final sparse linear aggregation of the rules through the estimate $\smash{\hat{\boldsymbol{\beta}}_{n,p_0}}$. To show that the stability converges to $1$, these estimates have to converge towards theoretical quantities that are independent of $\Dn$.
	Note that, throughout the paper, the final set of selected paths is denoted $\smash{\pathset_{M_n,n,p_{0}}}$. Here, for the sake of clarity, $\smash{\pathset_{M_n,n,p_{0}}}$ is now the post-treated set of paths extracted from the random forest, and $\smash{\pathset_{M_n,n,p_{0},\lambda}}$ the final set of selected paths in the ridge regression.
	
	\paragraph{$(i)$ Path extraction.}
	The first step of the proof is to show that the post-treated path extraction from the forest is consistent, i.e., in probability
	\begin{align} \label{extraction_consistency}
	\lim_{n \to \infty} \P(\pathset_{M_n,n,p_{0}} = \path^{\star}_{p_{0}}) = 1.
	\end{align}
	Using the continuous mapping theorem, it is easy to see that this result is a consequence of the consistency of  $\hat{p}_{M,n}(\path)$, i.e.,
	\[
	\lim\limits_{n \to \infty} \hat{p}_{M_{n},n}(\path) = p^{\star}(\path) \quad \textrm{in probability.}
	\]
	Since the output $Y$ is bounded (by Assumption (A2)), the consistency of $\hat{p}_{M,n}(\path)$ can be easily adapted from Theorem 1 of \citet{benard2020sirus} using Assumptions (A1) and (A2).
	Finally, the result still holds for the post-treated rule set because the post-treatment is a deterministic procedure.
	
	\paragraph{$(ii)$ Sparse linear aggregation.}
	Recall that the estimate $(\boldsymbol{\hat{\beta}}_{n,p_0}, \hat{\beta}_0)$ is defined as
	\begin{align} \label{def_beta}
	(\boldsymbol{\hat{\beta}}_{n,p_0}, \hat{\beta}_0) = \underset{\boldsymbol{\beta} \geq 0,\beta_0}{\textrm{argmin }} \ell_n(\boldsymbol{\beta},\beta_0),
	\end{align}
	where
	$\ell_n(\boldsymbol{\beta},\beta_0) =
	\frac{1}{n}||\mathbf{Y} - \beta_0 \boldsymbol{1_n} - \boldsymbol{\Gamma}_{n,p_0}\boldsymbol{\beta}||_2^2 + \lambda||\boldsymbol{\beta}||_2^2.$
	The dimension of $\beta$ is stochastic since it is equal to the number of extracted rules. To get rid of this technical issue in the following of the proof, we rewrite $\ell_n(\boldsymbol{\beta},\beta_0)$ to have $\boldsymbol{\beta}$ a parameter of fixed dimension $|\Pi|$, the total number of possible rules:
	\begin{align*}
	    \ell_n(\boldsymbol{\beta},\beta_0) =
	    \frac{1}{n} \sum_{i=1}^n \big(Y_i - \beta_0 - \sum_{\path \in \Pi} \beta_{\path} g_{n,\path}(\bX_i) \mathds{1}_{\path \in \pathset_{M_n,n,p_{0}}}\big)^2 + \lambda||\boldsymbol{\beta}||_2^2.
	\end{align*}
	By the law of large numbers and the previous result (\ref{extraction_consistency}), we have in probability
	\begin{align} \label{fct_consistency}
	\lim\limits_{n \to \infty} \ell_n(\boldsymbol{\beta},\beta_0) \nonumber =& \E\big[\big(Y - \beta_0 - \sum_{\path \in \path^{\star}_{p_{0}}} \beta_{\path} g_{\path}^{\star}(\bX)\big)^2\big] + \lambda||\boldsymbol{\beta}||_2^2 \stackrel{\textrm{def}}{=} \ell^{\star}(\boldsymbol{\beta},\beta_0),
	\end{align}
	where $g_{\path}^{\star}$ is the theoretical rule based on the path $\path$ and the theoretical quantiles.
	Since $Y$ is bounded, it is easy to see that each component of $\boldsymbol{\hat{\beta}}_{n,p_0}$ is bounded from the following inequalities:
	\begin{align*}
	\lambda||\boldsymbol{\hat{\beta}}_{n,p_0}||_2^2 \leq \ell_n(\boldsymbol{\hat{\beta}}_{n,p_0}, \hat{\beta}_0) \leq \ell_n(\boldsymbol{0},0) \leq \frac{||Y||_2^2}{n} \leq \underset{i}{\max} \hspace*{1mm} Y_i^2.
	\end{align*}
	Consequently, the optimization problem (\ref{def_beta}) can be equivalently written with $(\boldsymbol{\beta}, \beta_0)$ constrained to belong to a compact and convex set $K$. Since $\ell_n$ is convex and converges pointwise to $\ell^{\star}$ according to (\ref{fct_consistency}), the uniform convergence over the compact set $K$ also holds, i.e., in probability
	\begin{align}
	\lim \limits_{n \to \infty} \underset{(\boldsymbol{\beta}, \beta_0) \in K}{\sup} |\ell_n(\boldsymbol{\beta},\beta_0) - \ell^{\star}(\boldsymbol{\beta},\beta_0)| = 0.
	\end{align}
	Additionnally, since $\ell^{\star}$ is a quadratic convex function and the constraint domain $K$ is convex, $\ell^{\star}$ has a unique minimum that we denote $\boldsymbol{\beta}^{\star}_{p_0, \lambda}$.
	Finally, since the maximum of $\ell^{\star}$ is unique and $\ell_n$ uniformly converges to $\ell^{\star}$, we can apply theorem $5.7$ from \citet[page 45]{van2000asymptotic} to deduce that $(\boldsymbol{\hat{\beta}}_{n,p_0}, \hat{\beta}_0)$ is a consistent estimate of $\boldsymbol{\beta}^{\star}_{p_0, \lambda}$.
	We can conclude that, in probability,
	\begin{align*}
	\lim \limits_{n \to \infty} \P\big(\pathset_{M_n,n,p_{0},\lambda} =
	\{\path \in \path^{\star}_{p_{0}} : \beta^{\star}_{\path, p_0, \lambda} > 0 \}\big) = 1,
	\end{align*}
	and the final stability result follows from the continuous mapping theorem.
\end{proof}

\section{Computational Complexity}

The computational cost to fit SIRUS is similar to standard random forests, and its competitors: RuleFit, and Node harvest. The full tuning procedure costs about $10$ SIRUS fits.

\paragraph{SIRUS.}
SIRUS algorithm has several steps in its construction phase. We derive the computational complexity of each of them. Recall that $M$ is the number of trees, $p$ the number of input variables, and $n$ the sample size.
\begin{enumerate}

    \item \textbf{Forest growing: $O(M p n log(n))$}
    
    The forest growing is the most expensive step of SIRUS. The average computational complexity of a standard forest fit is $O(Mpnlog(n)^2)$ \citep{louppe2014understanding}. Since the depth of trees is fixed in SIRUS---see Section $3$, it reduces to $O(M p n log(n))$.
    
    A standard forest is grown so that its accuracy cannot be significantly improved with additional trees, which typically results in about $500$ trees.
    In SIRUS, the stopping criterion of the number of trees enforces that $95\%$ of the rules are identical over multiple runs with the same dataset (see Section \ref{sec_num_trees}). This is critical to have the forest structure converged and stabilize the final rule list. This leads to forests with a large number of trees, typically $10$ times the number for standard forests. On the other hand, shallow trees are grown and the computational complexity is proportional to the tree depth, which is about $log(n)$ for fully grown forests.
    
    Overall, the modified forest used in SIRUS is about the same computational cost as a standard forest, and has a slightly better computational complexity thanks to the fixed tree depth. 
    
    \item \textbf{Rule extraction: $O(M)$}
    
    Extracting the rules in a tree requires a number of operations proportional to the number of nodes, i.e. $O(1)$ since tree depth is fixed. With the appropriate data structure (a map), updating the forest count of the number of occurrences of the rules of a tree is also $O(1)$.
    Overall, the rule extraction is proportional to the number of trees in the forest, i.e., $O(M)$.
    
    \item \textbf{Rule post-treatment: $O(1)$}
    
    The post-treatment algorithm is only based on the rules and not on the sample. Since the number of extracted rules is bounded by a fixed limit of $25$, this step has a computational complexity of $O(1)$.
    
    \item \textbf{Rule aggregation: $O(n)$}
    
    Efficient algorithms \citep{friedman2010regularization} enable to fit a ridge regression and find the optimal penalization $\lambda$ with a linear complexity in the sample size $n$. In SIRUS, the predictors are the rules, whose number is upper bounded by $25$, and then the complexity of the rule aggregation is independent of $p$. Therefore the computational complexity of this step is $O(n)$. 
    
\end{enumerate}

Overall, the computational complexity of SIRUS is $O(Mpnlog(n))$, which is slightly better than standard random forests thanks to the use of shallow trees. Because of the large number of trees and the final ridge regression, the computational cost of SIRUS is comparable to standard forests in practice.

\paragraph{RuleFit/Node harvest Comparison.}
In both RuleFit and Node harvest, the first two steps of the procedure are also to grow a tree ensemble with limited tree depth and extract all possible rules. The complexity of this first phase is then similar to SIRUS: $O(Mpnlog(n))$. 
However, in the last step of the linear rule aggregation, all rules are combined in a sparse linear model, which is of linear complexity with $n$, but grows at faster rate than linear with the number of rules, i.e., the number of trees $M$ \citep{friedman2010regularization}.

As the tree ensemble growing is the computational costly step, SIRUS, RuleFit and Node harvest have a very comparable complexity. On one hand, SIRUS requires to grow more trees than its competitors. On the other hand, the final linear rule aggregation is done with few predictors in SIRUS, while it includes thousands of rules in RuleFit and Node harvest, which has a complexity faster than linear with $M$.

\paragraph{Tuning Procedure.}
The only parameter of SIRUS which requires fine tuning is $p_0$, which controls model sparsity. The optimal value is estimated by $10$-fold cross validation using a standard bi-objective optimization procedure to maximize both stability and predictivity.
For a fine grid of $p_0$ values, the unexplained variance and stability metric are computed for the associated SIRUS model through a cross-validation. Recall that the bounds of the $p_0$ grid are set to get the model size between $1$ and $25$ rules.
Next, we obtain a Pareto front, as illustrated in Figure \ref{fig_tuning}, where each point corresponds to a $p_0$ value of the tuning grid.
To find the optimal $p_0$, we compute the euclidean distance between each point and the ideal point of $0$ unexplained variance and $90\%$ stability. Notice that this ideal point is chosen for its empirical efficiency: the unexplained variance can be arbitrary close to $0$ depending on the data, whereas we do not observe a stability (with respect to data perturbation) higher than $90\%$ accross many datasets.
Finally, the optimal $p_0$ is the one minimizing the euclidean distance distance to the ideal point. Thus, the two objectives, stability and predictivity, are equally weighted. For a robust estimation of $p_0$, the cross-validation is repeated $10$ times and the median $p_0$ value is selected.
\begin{figure}[!ht]
	\begin{center}
		\includegraphics[height=7cm,width=7cm]{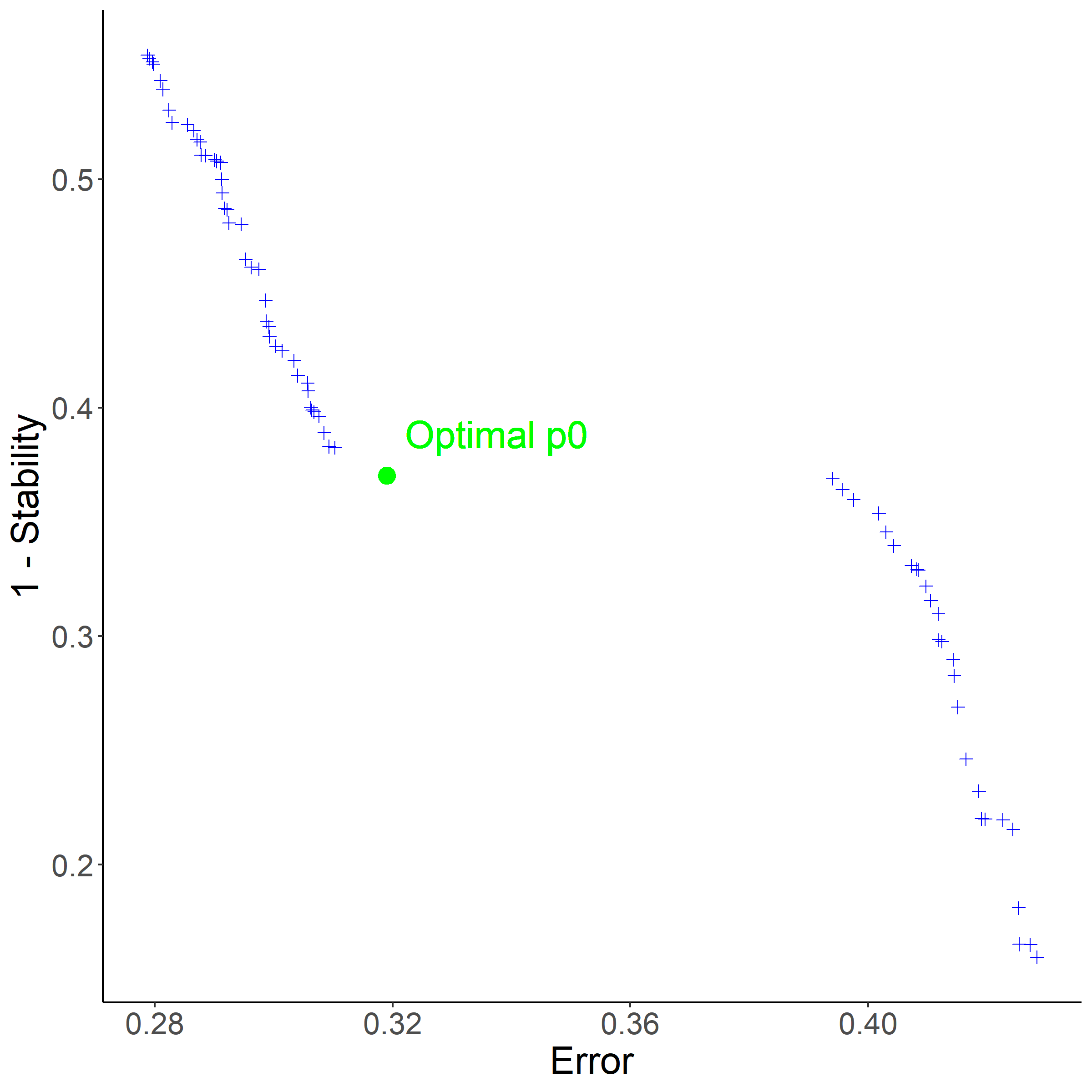}
		\caption{Pareto front of stability versus error (unexplained variance) when $p_0$ varies, with the optimal value in green for the ``Ozone'' dataset. The optimal point is the closest one to the ideal point $(0,0.1)$ of $0$ unexplained variance and $90\%$ stability.}
		\label{fig_tuning}
	\end{center}
\end{figure}

\paragraph{Tuning Complexity.}
The optimal $p_0$ value is estimated by a $10$-fold cross validation. The costly computational step of SIRUS is the forest growing. However, this step has to be done only once per fold. Then, $p_0$ can vary along a fine grid to extract more or less rules from each forest, and thus, get the accuracy associated to each $p_0$ at a total cost of about $10$ SIRUS fits.

\section{Random Forest Modifications}

As explained in Section 1 of the article, SIRUS uses random forests at its core. In order to stabilize the forest structure, we slightly modify the original algorithm from Breiman \citep{breiman2001random}: cut values at each tree node are limited to the $10$-empirical quantiles. In the first paragraph, we show how this restriction have a small impact on predictive accuracy, but is critical to stabilize the rule extraction. 
On the other hand, the rule selection mechanism naturally only keeps rules with one or two splits. Therefore, tree depth is fixed to $2$ to optimize the computational efficiency. In the second paragraph, this phenomenon is thoroughly explained.

\paragraph{Quantile discretization.}
In a typical setting where the number of predictors is $p = 100$, limiting cut values to the $10$-quantiles splits the input space in a fine grid of $10^{100}$ hyperrectangles. Therefore, restricting cuts to quantiles still leaves a high flexibility to the forest and enables to identify local patterns (it is still true in small dimension). To illustrate this, we run the following experiment: for each of the $8$ datasets, we compute the unexplained variance of respectively the standard forest and the forest where cuts are limited to the $10$-quantiles. Results are presented in Table \ref{table_RF_quantiles}, and we see that there is almost no decrease of accuracy except for one dataset.
Besides, notice that setting $q = n$ is equivalent as using original forests.
\begin{table}
    \setlength{\tabcolsep}{3pt}
	\centering
	\begin{tabular}{|c | c | c|}
		\hline \hline
		\textbf{Dataset} & \begin{tabular}{c}\textbf{Breiman Random} \\ \textbf{Forest} \end{tabular} & \begin{tabular}{c} \textbf{Random Forest}\\ \textbf{$10$-Quantile Cuts}\end{tabular}\\
		\hline
		Ozone & 0.25 (0.007) & 0.25 (0.006) \\	
		Mpg & 0.13 (0.003) & 0.13 (0.003)\\		
		Prostate & 0.46 (0.01) & 0.47 (0.02) \\	
		Housing & 0.13 (0.006) & 0.16 (0.004)\\	
		Diabetes & 0.55 (0.006) & 0.55 (0.007)\\	
		Machine & 0.13 (0.03) & 0.24 (0.02)\\	
		Abalone & 0.44 (0.002) & 0.49 (0.003)\\	
		Bones & 0.67 (0.01) & 0.68 (0.01)\\
		\hline \hline
	\end{tabular}
	\vspace*{1mm}
	\caption{Proportion of unexplained variance (estimated over a $10$-fold cross-validation) for various public datasets to compare two algorithms: Breiman's random forest and the forest where split values are limited to the $10$-empirical quantiles. Standard deviations are computed over multiple repetitions of the cross-validation and displayed in brackets.}
	\label{table_RF_quantiles}
\end{table}

On the other hand, such discretization is critical for the stability of the rule
selection.
Recall that the importance of each rule $\hat{p}_{M,n}(\path)$ is defined as the proportion of trees which contain its associated path $\path$, and that the rule selection is based on $\hat{p}_{M,n}(\path) > p_0$. In the forest growing, data is bootstrapped prior to the construction of each tree. Without the quantile discretization, this data perturbation results in small variation between the cut values across different nodes, and then the dilution of $\hat{p}_{M,n}(\path)$ between highly similar rules. Thus, the rule selection procedure becomes inefficient.
More formally, $\hat{p}_{M,n}(\path)$ is defined by
\[
\hat{p}_{M,n}(\path)=\frac{1}{M}\sum_{\ell=1}^{M}\mathds{1}_{\path\in T(\Theta_{\ell},\mathscr{D}_{n})},
\]
where $T(\Theta_{\ell},\mathscr{D}_{n})$ is the list of paths extracted from the $\ell$-th tree of the forest.
The expected value of the importance of a given rule is
\begin{align*}
    \E[\hat{p}_{M,n}(\path)] = \frac{1}{M}\sum_{\ell=1}^{M} \E[\mathds{1}_{\path\in T(\Theta_{\ell},\mathscr{D}_{n})}]
    = \P(\path\in T(\Theta,\mathscr{D}_{n})).
\end{align*}
Without the discretization, $T(\Theta,\mathscr{D}_{n})$ is a random set that takes value in an uncountable space, and consequently
\begin{align*}
    \E[\hat{p}_{M,n}(\path)]
    = \P(\path\in T(\Theta_{\ell},\mathscr{D}_{n})) = 0,
\end{align*}
and all rules are equally not important in average.
In practice, since $\Dn$ is of finite size and the random forest cuts at mid distance between two points, it is still possible to compute $\hat{p}_{M,n}(\path)$ and select rules for a given dataset. However, such procedure is highly unstable with respect to data perturbation since we have $\E[\hat{p}_{M,n}(\path)] = 0$ for all possible paths.

\paragraph{Tree depth.}
When SIRUS is fit using fully grown trees, the final set of rules $\pathset_{M,n,p_0}$ contains almost exclusively rules made of one or two splits, and very rarely of three splits.
Although this may appear surprising at first glance, this phenomenon is in fact expected. 
Indeed, rules made of multiple splits are extracted from deeper tree levels and are thus more sensitive to data perturbation by construction. This results in much smaller values of $\hat{p}_{M,n}(\path)$ for rules with a high number of splits, and then deletion from the final set of path through the threshold $p_0$: $\smash{\pathset_{M,n,p_{0}}=\{ \path \in \Pi:\hat{p}_{M,n}(\path)>p_{0}\}}$. To illustrate this, let us consider the following typical example with $p = 100$ input variables and $q = 10$ quantiles. There are $2 q p = 2 \times 100 \times 10 = 2\times 10^3$ distinct rules of one split, about $(2 q p)^2 \approx 10^6$ distinct rules of two splits, and about $(2 q p)^3 \approx 10^{10}$ distinct rules of three splits. Using only rules of one split is too restrictive since it generates a small model class (a thousand rules for $100$ input variables) and does not handle variable interactions. On the other hand, rules of two splits are numerous (a million) and thus provide a large flexibility to SIRUS. More importantly, since there are $10$ billion rules of three splits, a stable selection of a few of them is clearly an impossible task, and such complex rules are naturally discarded by SIRUS. 

In SIRUS, tree depth is set to $2$ to reduce the computational cost while leaving the output list of rules untouched as previously explained. We augment the experiments of Section $3$ of the article with an additional column in Table \ref{table_error_2obj}: ``\textbf{SIRUS $50$ rules \& d$=3$}''.
Recall that, in the column ``\textbf{SIRUS $50$ rules}'', $p_0$ is set manually to extract $100$ rules from the forest leading to final lists of about $50$ rules (similar size as RuleFit and Node harvest models), an improved accuracy (reaching RuleFit performance), while stability drops to around $50\%$ ($70-80\%$ when $p_0$ is tuned). In the last column, tree depth is set to $3$ with the same augmented model size. We observe no accuracy improvement over a tree depth of $2$.

This analysis of tree depth is not new. Indeed, both RuleFit \citep{friedman2008predictive} and Node harvest \citep{meinshausen2010node} articles discuss the optimal tree depth for the rule extraction from a tree ensemble in their experiments.
They both conclude that the optimal depth is $2$. Hence, the same hard limit of $2$ is used in Node harvest. RuleFit is slightly less restrictive: for each tree, its depth is randomly sampled with an exponential distribution concentrated on $2$, but allowing few trees of depth $1$, $3$ and $4$. We insist that they both reach such conclusion without considering stability issues, but only focusing on accuracy.
\begin{table}
    \renewcommand\thetable{3}
    \setlength{\tabcolsep}{3pt}
	\centering
	\begin{tabular}{|c | c | c | c | c | c | c | c | c | c |}
		\hline \hline
		\textbf{Dataset} & \begin{tabular}{c}\textbf{Random} \\ \textbf{Forest} \end{tabular} & \textbf{CART} & \textbf{RuleFit} & \begin{tabular}{c}\textbf{Node} \\ \textbf{harvest}\end{tabular} & \textbf{SIRUS} & \begin{tabular}{c} \textbf{SIRUS} \\ \textbf{sparse} \end{tabular} & \begin{tabular}{c}\textbf{SIRUS} \\ \textbf{50 rules}\end{tabular} & \begin{tabular}{c}\textbf{SIRUS} \\ \textbf{50 rules \& d=3}\end{tabular}\\ 
		\hline
		Ozone & 0.25 & 0.36 & \textbf{0.27} & 0.31 & 0.32 & 0.32 & \textbf{0.26} & \textbf{0.27} \\	
		Mpg & 0.13 & 0.20 & \textbf{0.15} & 0.20 & 0.20 & 0.20 & \textbf{0.15} & \textbf{0.15} \\		
		Prostate & 0.48 & 0.60 & \textbf{0.53} & \textbf{0.52} & \textbf{0.55} & 0.51 & \textbf{0.54} & \textbf{0.55}\\	
		Housing & 0.13 & 0.28 & \textbf{0.16}  & 0.24 & 0.30 & 0.31 & 0.20 & 0.21 \\	
		Diabetes & 0.55 & 0.67 & \textbf{0.55} & \textbf{0.58} & \textbf{0.56} & 0.56 & \textbf{0.55} & \textbf{0.55} \\	
		Machine & 0.13 & 0.39 & \textbf{0.26} & \textbf{0.29} & \textbf{0.29} & 0.32 & \textbf{0.27} & \textbf{0.26} \\	
		Abalone & 0.44 & 0.56 & \textbf{0.46} & 0.61 & 0.66 & 0.64 & 0.64 & 0.65 \\	
		Bones & 0.67 & 0.67 & \textbf{0.70} & \textbf{0.70} & \textbf{0.73} & 0.77 & \textbf{0.73} & \textbf{0.75} \\
		\hline \hline
	\end{tabular}
	\vspace*{1mm}
	\caption{Proportion of unexplained variance estimated over a $10$-fold cross-validation for various public datasets. For rule algorithms only, i.e., RuleFit, Node harvest, and SIRUS, minimum values are displayed in bold, as well as values within 10\% of the minimum for each dataset (``SIRUS sparse'' put aside).}
	\label{table_error_2obj}
\end{table}

\section{Post-treatment Illustration}

We illustrate the post-treatment procedure with the ``Machine'' dataset.
Table \ref{fig_machine_raw} provides the initial raw list of $17$ rules on the left, and the final post-treated $9$-rule list on the right, using $p_0 = 0.072$. The rules removed from the raw list are highlighted in red and orange. Red rules have one constraint and are identical to a previous rule with the constraint sign reversed. Notice that two such rules (e.g. rules $1$ and $2$) correspond to the left and right child nodes at the first level of a tree. Thus, they belong to the same trees of the forest and their associated occurrence frequencies $\hat{p}(\path)$ are equal. We always keep the rule with the sign ``<'': this choice is somewhat arbitrary and of minor importance since the two rules are identical. Orange rules have two constraints and are linearly dependent on other previous rules. For example for rule $12$, there exist $3$ real numbers $\alpha_1$, $\alpha_5$, and $\alpha_7$ such that, for all $\bx \in \R^p$
\begin{align*}
    g_{\path_{12}}(\bx) = \alpha_1 g_{\path_1}(\bx) + \alpha_5 g_{\path_5}(\bx) + \alpha_7 g_{\path_7}(\bx).
\end{align*}
Observe that rules $12$ and $7$ involve the same variables and thresholds, but one of the sign constraints is reversed. The estimated rule outputs $\hat{Y}$ are of course different between rules $12$ and $7$ because they identify two different quarters of the input space. The outputs of rule $7$ have a wider gap than the ones of rule $12$, and consequently the CART-splitting criterion of rule $12$ is smaller, which also implies a smaller occurrence frequency, i.e., $\hat{p}(\path_{12}) < \hat{p}(\path_{7})$. Therefore rule $12$ is removed rather than rule $7$. The same reasoning applies to rules $15$ and $17$.
\begin{table}
    \small
	\begin{center}
		\setlength{\fboxrule}{1.5pt}
		\fbox{\begin{minipage}{0.47\textwidth}
				\setlength{\tabcolsep}{0.6pt}
				\begin{tabular}{c | c c c c c c }
					1 & \hspace*{1mm} \textbf{if } & $\textrm{MMAX} < 32000$ & \textbf{ then } & $\hat{Y} = 61$ & \textbf{ else } & $\hat{Y} = 408$ \\[1mm]
					\rowcolor{red!60}
					2 & \hspace*{1mm} \textbf{if } & $\textrm{MMAX} \geq 32000$ & \textbf{ then } & $\hat{Y} = 408$ & \textbf{ else } & $\hat{Y} = 61$ \\[1mm]
					3 & \hspace*{1mm} \textbf{if } & $\textrm{MMIN} < 8000$ & \textbf{ then } & $\hat{Y} = 62$ & \textbf{ else } & $\hat{Y} = 386$ \\[1mm]
					\rowcolor{red!60}
					4 & \hspace*{1mm} \textbf{if } & $\textrm{MMIN} \geq 8000$ & \textbf{ then } & $\hat{Y} = 386$ & \textbf{ else } & $\hat{Y} = 62$ \\[1mm]
					5 & \hspace*{1mm} \textbf{if } & $\textrm{CACH} < 64$ & \textbf{ then } & $\hat{Y} = 56$ & \textbf{ else } & $\hat{Y} = 334$ \\[1mm]
					\rowcolor{red!60}
					6 & \hspace*{1mm} \textbf{if } & $\textrm{CACH} \geq 64$ & \textbf{ then } & $\hat{Y} = 334$ & \textbf{ else } & $\hat{Y} = 56$ \\[1mm]
					7 & \hspace*{1mm} \textbf{if } & $\left\{\begin{tabular}{c} $\textrm{MMAX} \geq 32000$ \\ \textbf{\&} $\textrm{CACH} \geq 64$ \end{tabular}\right.$ & \textbf{ then } & $\hat{Y} = 517$ & \textbf{ else } &  $\hat{Y} = 67$  \\[1mm]	
					8 & \hspace*{1mm} \textbf{if } & $\textrm{CHMIN} < 8$ & \textbf{ then } & $\hat{Y} = 50$ & \textbf{ else } & $\hat{Y} = 312$ \\[1mm]
					\rowcolor{red!60}
					9 & \hspace*{1mm} \textbf{if } & $\textrm{CHMIN} \geq 8$ & \textbf{ then } & $\hat{Y} = 312$ & \textbf{ else } & $\hat{Y} = 50$ \\[1mm]
					10 & \hspace*{1mm} \textbf{if } & $\textrm{MYCT} < 50$ & \textbf{ then } & $\hat{Y} = 335$ & \textbf{ else } & $\hat{Y} = 58$ \\[1mm]
					\rowcolor{red!60}
					11 & \hspace*{1mm} \textbf{if } & $\textrm{MYCT} \geq 50$ & \textbf{ then } & $\hat{Y} = 58$ & \textbf{ else } & $\hat{Y} = 335$ \\[1mm]
					\rowcolor{orange!60}
                    12 & \hspace*{1mm} \textbf{if } & $\left\{\begin{tabular}{c} $\textrm{MMAX} \geq 32000$ \\ \textbf{\&} $\textrm{CACH} < 64$ \end{tabular}\right.$ & \textbf{ then } & $\hat{Y} = 192$ & \textbf{ else } &  $\hat{Y} = 102$  \\[3mm]
                    13 & \hspace*{1mm} \textbf{if } & $\left\{\begin{tabular}{c} $\textrm{MMAX} < 32000$ \\ \textbf{\&} $\textrm{CHMIN} \geq 8$ \end{tabular}\right.$ & \textbf{ then } & $\hat{Y} = 157$ & \textbf{ else } &  $\hat{Y} = 100$  \\[3mm]
                    14 & \hspace*{1mm} \textbf{if } & $\left\{\begin{tabular}{c} $\textrm{MMAX} < 32000$ \\ \textbf{\&} $\textrm{CHMIN} \geq 12$ \end{tabular}\right.$ & \textbf{ then } & $\hat{Y} = 554$ & \textbf{ else } &  $\hat{Y} = 73$  \\[3mm]
                    \rowcolor{orange!60}
                    15 & \hspace*{1mm} \textbf{if } & $\left\{\begin{tabular}{c} $\textrm{MMAX} \geq 32000$ \\ \textbf{\&} $\textrm{CHMIN} < 12$ \end{tabular}\right.$ & \textbf{ then } & $\hat{Y} = 252$ & \textbf{ else } &  $\hat{Y} = 96$  \\[3mm]
                    16 & \hspace*{1mm} \textbf{if } & $\left\{\begin{tabular}{c} $\textrm{MMIN} \geq 8000$ \\ \textbf{\&} $\textrm{CHMIN} \geq 12$ \end{tabular}\right.$ & \textbf{ then } & $\hat{Y} = 586$ & \textbf{ else } &  $\hat{Y} = 76$  \\[3mm]
                    \rowcolor{orange!60}
                    17 & \hspace*{1mm} \textbf{if } & $\left\{\begin{tabular}{c} $\textrm{MMIN} \geq 8000$ \\ \textbf{\&} $\textrm{CHMIN} < 12$ \end{tabular}\right.$ & \textbf{ then } & $\hat{Y} = 236$ & \textbf{ else } &  $\hat{Y} = 94$ 
                \end{tabular}
		\end{minipage}}
		\setlength{\fboxrule}{1.5pt}
		\fbox{\begin{minipage}{0.47\textwidth}
				\setlength{\tabcolsep}{0.6pt}
				\begin{tabular}{c | c c c c c c }
					1 & \hspace*{1mm} \textbf{if } & $\textrm{MMAX} < 32000$ & \textbf{ then } & $\hat{Y} = 61$ & \textbf{ else } & $\hat{Y} = 408$ \\[1mm]
					3 & \hspace*{1mm} \textbf{if } & $\textrm{MMIN} < 8000$ & \textbf{ then } & $\hat{Y} = 62$ & \textbf{ else } & $\hat{Y} = 386$ \\[1mm]
					5 & \hspace*{1mm} \textbf{if } & $\textrm{CACH} < 64$ & \textbf{ then } & $\hat{Y} = 56$ & \textbf{ else } & $\hat{Y} = 334$ \\[1mm]
					7 & \hspace*{1mm} \textbf{if } & $\left\{\begin{tabular}{c} $\textrm{MMAX} \geq 32000$ \\ \textbf{\&} $\textrm{CACH} \geq 64$ \end{tabular}\right.$ & \textbf{ then } & $\hat{Y} = 517$ & \textbf{ else } &  $\hat{Y} = 67$  \\[1mm]	
					8 & \hspace*{1mm} \textbf{if } & $\textrm{CHMIN} < 8$ & \textbf{ then } & $\hat{Y} = 50$ & \textbf{ else } & $\hat{Y} = 312$ \\[1mm]
					10 & \hspace*{1mm} \textbf{if } & $\textrm{MYCT} < 50$ & \textbf{ then } & $\hat{Y} = 335$ & \textbf{ else } & $\hat{Y} = 58$ \\[1mm]
                    13 & \hspace*{1mm} \textbf{if } & $\left\{\begin{tabular}{c} $\textrm{MMAX} < 32000$ \\ \textbf{\&} $\textrm{CHMIN} \geq 8$ \end{tabular}\right.$ & \textbf{ then } & $\hat{Y} = 157$ & \textbf{ else } &  $\hat{Y} = 100$  \\[3mm]
                    14 & \hspace*{1mm} \textbf{if } & $\left\{\begin{tabular}{c} $\textrm{MMAX} < 32000$ \\ \textbf{\&} $\textrm{CHMIN} \geq 12$ \end{tabular}\right.$ & \textbf{ then } & $\hat{Y} = 554$ & \textbf{ else } &  $\hat{Y} = 73$  \\[3mm]
                    16 & \hspace*{1mm} \textbf{if } & $\left\{\begin{tabular}{c} $\textrm{MMIN} \geq 8000$ \\ \textbf{\&} $\textrm{CHMIN} \geq 12$ \end{tabular}\right.$ & \textbf{ then } & $\hat{Y} = 586$ & \textbf{ else } &  $\hat{Y} = 76$  \\[3mm]
                \end{tabular}
		\end{minipage}}
	\end{center}
	\caption{\small{SIRUS post-treatment of the extracted raw list of rules for the ``Machine'' dataset: the raw list of rules on the left, and the final post-treated rule list on the right (removed rules are highlighted in red for one constraint rules and in orange for two constraint rules).}}
	\label{fig_machine_raw}
\end{table}

\section{Rule Format}
The format of the rules with an else clause for the uncovered data points differs from the standard format in the rule learning literature. Indeed, in classical algorithms, a prediction is generated for a given query point by aggregating the outputs of the rules satisfied by the point. A default rule usually provides predictions for all query points which satisfy no rule. 
First, observe that the intercept in the final linear aggregation of rules in SIRUS can play the role of a default rule. Secondly, removing the else clause of the rules selected by SIRUS results in an equivalent formulation of the linear regression problem up to the intercept.
More importantly, the format with an else clause is required for the stability and modularity \citep{murdoch2019interpretable} properties of SIRUS.
\paragraph{Equivalent Formulation.}
Rules are originally defined in SIRUS as
\begin{align*}
\hat{g}_{n,\path} (\bx) = 
\begin{cases}
\bar{Y}_{\path}^{(1)} &\textrm{if}~\bx \in \path \\
\bar{Y}_{\path}^{(0)} &\textrm{otherwise,}
\end{cases}
\end{align*}
where $\textrm{if}~\bx \in \path$ indicates whether the query point $\bx$ satisfies the rule associated with path $\path$ or not, $\smash{\bar{Y}_{\path}^{(1)}}$ is the output average of the training points which satisfy the rule, and symmetrically $\smash{\bar{Y}_{\path}^{(0)}}$ is the output average of the training point not covered by the rule.
The original linear aggregation of the rules is
\begin{align*}
\hat{m}_{M,n,p_{0}}(\bx)
= \hat{\beta}_0 + \sum_{\path \in \pathset_{M,n,p_{0}}} \hat{\beta}_{n, \path} \hat{g}_{n, \path}( \bx).
\end{align*}
Now we define the rules without the else clause by $\hat{h}_{n,\path} (\bx) = (\bar{Y}_{\path}^{(1)} - \bar{Y}_{\path}^{(0)}) \mathds{1}_{\bx \in \path}$, and we can rewrite SIRUS estimate as
\begin{align*}
\hat{m}_{M,n,p_{0}}(\bx)
=& \big(\hat{\beta}_0 + \sum_{\path \in \pathset_{M,n,p_{0}}} \hat{\beta}_{n, \path} \bar{Y}_{\path}^{(0)}\big) + \sum_{\path \in \pathset_{M,n,p_{0}}} \hat{\beta}_{n, \path} \hat{h}_{n, \path}( \bx) \\
=& \tilde{\beta}_0 + \sum_{\path \in \pathset_{M,n,p_{0}}} \hat{\beta}_{n, \path} \hat{h}_{n, \path}( \bx).
\end{align*}
Therefore the two models with or without the else clause are equivalent up to the intercept.

\paragraph{Stability.} The problem of defining rules without the else clause lies in the rule selection. Indeed, rules associated with left ($L$) and right ($R$) nodes at the first level of a tree are identical:
\begin{align*}
\hat{g}_{n,L} (\bx) = \hat{g}_{n,R} (\bx) = \bar{Y}_{L}\mathds{1}_{\bx \in L} + \bar{Y}_{R} \mathds{1}_{\bx \in R}.
\end{align*}
Without the else clause, these two rules become different estimates:
\begin{align*}
\hat{h}_{n,L} (\bx) = (\bar{Y}_{L} - \bar{Y}_{R}) \mathds{1}_{\bx \in L}, \\
\hat{h}_{n,R} (\bx) = (\bar{Y}_{R} - \bar{Y}_{L}) \mathds{1}_{\bx \in R}.
\end{align*}
However, $\hat{h}_{n,L}$ and $\hat{h}_{n,R}$ are linearly dependent, since 
$\hat{h}_{n,L} (\bx) - \hat{h}_{n,R} (\bx) = \bar{Y}_{L} - \bar{Y}_{R}$, which does not depend on the query point $\bx$. This linear dependence between predictors makes the linear aggregation of the rules ill-defined. One of two rule could be removed randomly, but this would strongly hurt stability. 

\paragraph{Modularity.} 
\citet{murdoch2019interpretable} specify different properties to assess model simplicity: sparsity, simulatability, and modularity. A model is sparse when it uses only a small fraction of the input variables, e.g. the lasso. A model is simulatable if it is possible for humans to perform predictions by hands, e.g. shallow decision trees. A model is modular when it is possible to analyze a meaningful portion of it alone. Typically, rule models are modular since one can analyze the rules one by one. In that case, the average of the output values for instances not covered by the rule is an interesting insight.

\section{Dataset Descriptions} \label{appendix_data}
\begin{table}[H]
	\setlength{\tabcolsep}{2pt}
	\centering
	\begin{tabular}{|c | c | c | c|}
		\hline \hline
		\textbf{Dataset} & \textbf{Sample Size} & \begin{tabular}{c}\textbf{Total Number} \\ \textbf{of Variables} \end{tabular} &
		\begin{tabular}{c}\textbf{Number of} \\ \textbf{Categorical} \\ \textbf{Variables} \end{tabular} \\
		\hline
		Ozone & 330 & 9 & 0 \\	
		Mpg & 398 & 7 & 0 \\	
		Prostate & 97 & 8 & 0 \\	
		Housing & 506 & 13 & 0 \\	
		Diabetes & 442 & 10 & 0 \\	
		Machine & 209 & 6 & 0 \\	
		Abalone & 4177 & 8 & 1 \\	
		Bones & 485 & 3 & 2 \\
		\hline \hline
	\end{tabular}
	\vspace*{1.5mm}
	\caption{\small{Description of datasets}}
	\label{table_datasets}
\end{table}

\section{Number of Trees} \label{sec_num_trees}
The stability, predictivity, and computation time of SIRUS increase with the number of trees. Thus a stopping criterion is designed to grow the minimum number of trees that ensures stability and predictivity to be close to their maximum. It happens in practice that stabilizing the rule list is computationally more demanding in the number of trees than reaching a high predictivity. Therefore the stopping criterion is only based on stability, and defined as the minimum number of trees such that when SIRUS is fit twice on the same given dataset, $95\%$ of the rules are shared by the two models in average. 

To this aim, we introduce $1 - \varepsilon_{M,n,p_0}$, an estimate of the mean stability $\E[\hat{S}_{M_n,n,p_0}|\Dn]$ when SIRUS is fit twice on the same dataset $\Dn$. $\varepsilon_{M,n,p_0}$ is defined by
\begin{align*}
\varepsilon_{M,n,p_0} = \frac{\sum_{\path \in \Pi} z_{M,n,p_0}(\path) ( 1 - z_{M,n,p_0}(\path))}{\sum_{\path \in \Pi} ( 1 - z_{M,n,p_0}(\path)) },
\end{align*}
where $z_{M,n,p_0}(\path) = \Phi( M p_{0}, M, p_n(\path))$, the cdf of a binomial distribution with parameter $p_n(\path)=\E[\hat{p}_{M_{n},n}(\path)|\Dn]$, $M$ trials, evaluated at $M p_{0}$.
It happens that $\varepsilon_{M,n,p_0}$ is quite insensitive to $p_0$. Consequently it is simply averaged over a grid  $\hat{V}_{M,n}$ of many possible values of $p_0$. 
Therefore, the number of trees is set, for $\alpha = 0.05$, by
\begin{align*} \label{criterion_M}
\underset{M}{\textrm{argmin}} \Big\{ \frac{1}{| \hat{V}_{M,n}|} \sum_{p_0 \in \hat{V}_{M,n}}
\varepsilon_{M,n,p_0} < \alpha \Big\},
\end{align*}
to ensure that $95\%$ of the rules are shared by the two models in average.
See Section $4$ from \citet{benard2020sirus} for a thorough explanation of this stopping criterion.

\end{document}